\documentclass{article}
\usepackage[colorlinks,urlcolor=blue,linkcolor=blue,citecolor=blue]{hyperref}

\usepackage{color,array}
\usepackage{cite}
\usepackage[caption=false,font=footnotesize]{subfig}
\usepackage{bm}
\usepackage{amsmath,amssymb,amsfonts}
\usepackage{algorithmic}
\usepackage{graphicx}
\usepackage{textcomp}
\usepackage{xcolor}
\usepackage{algorithm,algorithmic}

\usepackage{epstopdf}
\usepackage{amsthm}
\usepackage{lineno}
\usepackage{mathrsfs}
\usepackage{caption}
\usepackage{cases}
\newtheorem{definition}{Definition}

\newtheorem{theorem}{Theorem}%[section]
\DeclareMathOperator*{\argmin}{argmin}

\begin{document}

\title{Tucker-O-Minus Decomposition for Multi-view Tensor Subspace Clustering}

%\author{Yingcong~Lu, Yipeng~Liu, ~\IEEEmembership{Senior Member,~IEEE}, Zhen Long, Zhangxin Chen, Ce~Zhu, \IEEEmembership{Fellow, IEEE}
\author{Yingcong~Lu, Yipeng~Liu, Zhen Long, Zhangxin Chen, Ce~Zhu
%\thanks{This work was supported by the National Natural Science Foundation of China (NSFC) under Grant 62171088, Grant 6220106011 and Grant U19A2052.}
\thanks{All the authors are with the School of Information and Communication Engineering, University of Electronic Science and Technology of China (UESTC), Chengdu, 611731, China. E-mail: yipengliu@uestc.edu.cn}
}

%\markboth{IEEE Transactions on Artificial Intelligence, Vol. 00, No. 00,  Year}
%{First A. Author \MakeLowercase{\textit{et al.}}: Bare Demo of IEEEtai.cls for IEEE Journals of IEEE Transactions on Artificial Intelligence}

\maketitle

\begin{abstract}
With powerful ability to exploit latent structure of self-representation information, different tensor decompositions have been employed into low rank multi-view clustering (LRMVC) models for achieving significant performance. However, current approaches suffer from a series of problems related to those tensor decomposition, such as the unbalanced matricization scheme, rotation sensitivity, deficient correlations capture and so forth. All these will lead to LRMVC having insufficient access to global information, which is contrary to the target of multi-view clustering. To alleviate these problems, we propose a new tensor decomposition called Tucker-O-Minus Decomposition (TOMD) for multi-view clustering. Specifically, based on the Tucker format, we additionally employ the O-minus structure, which consists of a circle with an efficient bridge linking two weekly correlated factors. In this way, the core tensor in Tucker format is replaced by the O-minus architecture with a more balanced structure, and the enhanced capacity of capturing the global low rank information will be achieved. The proposed TOMD also provides more compact and powerful representation abilities for the self-representation tensor, simultaneously. The alternating direction method of multipliers is used to solve the proposed model TOMD-MVC. Numerical experiments on six benchmark data sets demonstrate the superiority of our proposed method in terms of F-score, precision, recall, normalized mutual information, adjusted rand index, and accuracy.
\end{abstract}

\section{Introduction}

%\IEEEPARstart{T}{ensor}, a higher order generation of vector and matrix, provides a natural representation for high order data. For example, ORL multi-view data set~\cite{zhang2020tensorized} is a 3rd-order tensor $\mathcal{X}^{I \times C_{v} \times V}$ $(v = 1,\cdots,V)$, where $I$ is the number of samples, $C_{v}$ is the feature size of the $v$-th view, and $V$ is the number of views. Tensor decomposition allows us to simultaneously explore all dimensions to obtain more latent information, which has attracted attention in a series of fields, e.g., image processing~\cite{chen2021hierarchical,long2019low}, machine learning~\cite{ji2019survey,kaliyar2021deepfake} and signal processing~\cite{chen2021tensor,adali2022reproducibility}. Tensor-based multi-view clustering (MVC) is one of them, which separates multi-view data into clusters by exploiting their latent information. 
Tensor, a higher order generation of vector and matrix, provides a natural representation for high order data. For example, ORL multi-view data set~\cite{zhang2020tensorized} is a 3rd-order tensor $\mathcal{X}^{I \times C_{v} \times V}$ $(v = 1,\cdots,V)$, where $I$ is the number of samples, $C_{v}$ is the feature size of the $v$-th view, and $V$ is the number of views. Tensor decomposition allows us to simultaneously explore all dimensions to obtain more latent information, which has attracted attention in a series of fields, e.g., image processing~\cite{chen2021hierarchical,long2019low}, machine learning~\cite{ji2019survey,kaliyar2021deepfake} and signal processing~\cite{chen2021tensor,adali2022reproducibility}. Tensor-based multi-view clustering (MVC) is one of them, which separates multi-view data into clusters by exploiting their latent information. 

Most tensor MVC methods are based on the assumption that their self-representation tensors are low rank~\cite{zhang2015low}. For example, Chen et al. \cite{chen2021low} combine the low-rank tensor graph and the subspace clustering into a unified model in multi-view clustering, which applies the Tucker decomposition \cite{tucker1966some,tucker1964extension} to explore the low-rank information of representation tensor. With the emergence of a new type of decomposition for 3rd-order tensors \cite{kilmer2008third,kilmer2011factorization}, multiple multi-view subspace clustering methods based on the tensor singular value decomposition (t-SVD) have been proposed \cite{wu2019essential,wu2020unified,zheng2022multi}. In addition, tensor networks provide a more compact and flexible representation for higher order tensor than traditional tensor decomposition~\cite{liu2020smooth,liu2021tensor,liu2020generalizing,long2021bayesian}.  In this way, Yu et al. \cite{yu2020graph} have proposed a novel non-negative tensor ring (NTR) decomposition and graph-regularized NTR (GNTR) for the non-negative multi-way representation learning with satisfying performance in clustering tasks. 

%Although the aforementioned methods have achieved promising clustering performance, there are still several problems. Since the Tucker rank is related to the unbalanced mode-$n$ unfolding matrices~\cite{bengua2017efficient}, capturing the global information of the self-representation tensor by simply processing Tucker decomposition may be difficult in LRMVC. Besides, the t-SVD suffers from rotation sensitivity and the low rank information can not be fully discovered in the 3rd mode~\cite{feng2021multi}. Thus, the t-SVD based methods always need to utilize the rotation operation of the self-representation tensor to explore the correlations across different views. Furthermore, tensor ring (TR) has shown better performance in exploring the low-rank information with the well-balanced matricization scheme. However, the interaction of neighboring modes in TR is stronger than that of two modes separated by a large distance, whose correlations have been ignored. 
Although the aforementioned methods have achieved promising clustering performance, there are still several problems. Since the Tucker rank is related to the unbalanced mode-$n$ unfolding matrices~\cite{bengua2017efficient}, capturing the global information of the self-representation tensor by simply processing Tucker decomposition may be difficult in LRMVC. Besides, the t-SVD suffers from rotation sensitivity and the low rank information can not be fully discovered in the 3rd mode~\cite{feng2021multi}. Thus, the t-SVD based methods always need to utilize the rotation operation of the self-representation tensor to explore the correlations across different views. Naturally, the acquisition of the correlations among samples will be inadequate. Furthermore, tensor ring (TR) has shown better performance in exploring the low-rank information with the well-balanced matricization scheme. However, the interaction of neighboring modes in TR is stronger than that of two modes separated by a large distance, whose correlations are also have been ignored. 

\begin{figure}[htbp]
\centering
\includegraphics[scale=.15]{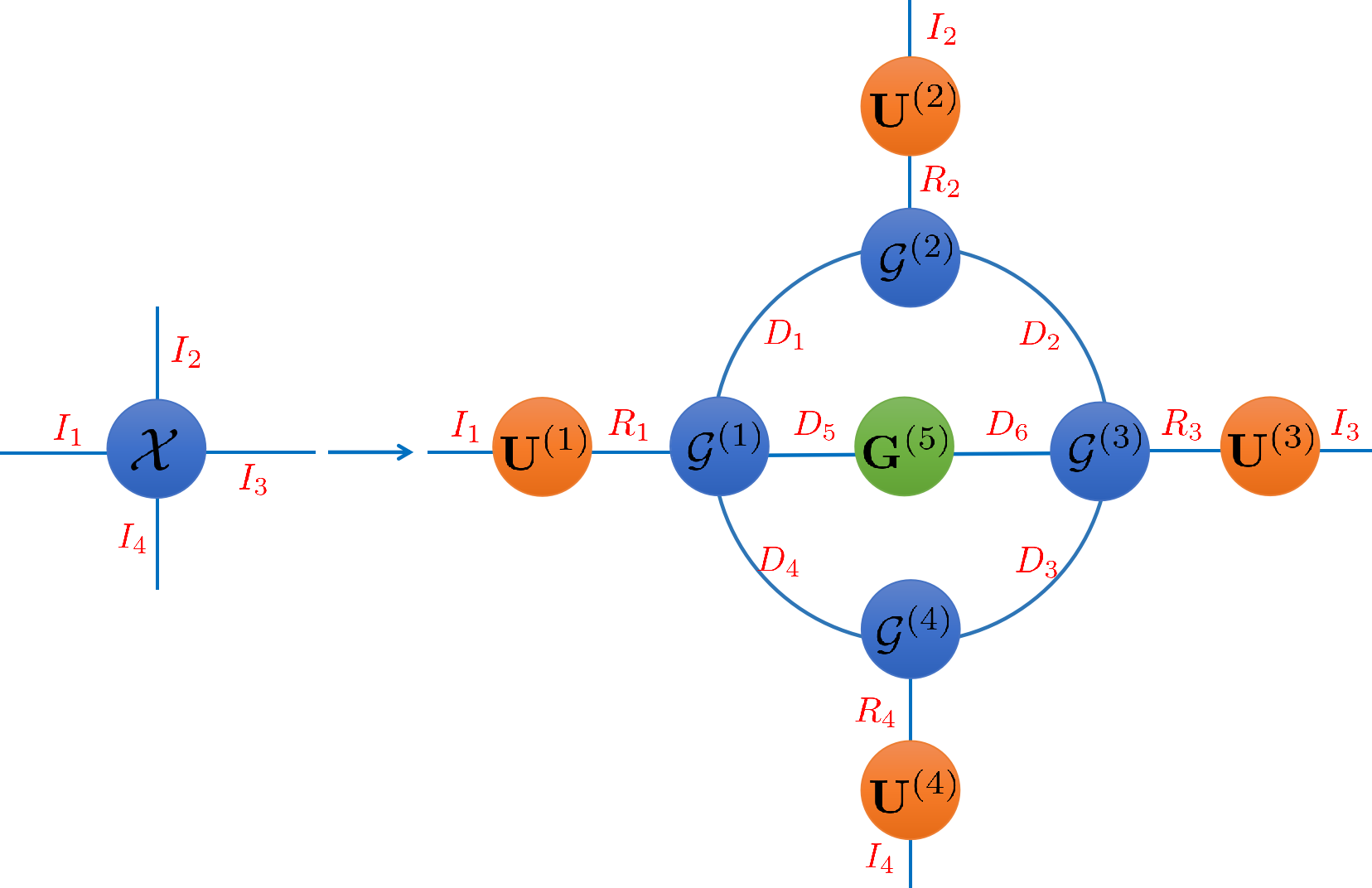}
\caption{The graphical illustration of TOMD for a 4th-order tensor $\mathcal{X}\in\mathbb{R}^{I_1 \times I_2 \times I_3 \times I_4}$, where $\mathbf{U}^{(i)}\in\mathbb{R}^{I_{i}\times R_{i}}$ $(i=1,\cdots,4)$ are factor matrices,  $\mathcal{G}^{(n)}$, $n=1,\cdots,5$, are the sub-tensors.}
\label{fig: ndprocess}       % Give a unique label
\end{figure}

Considering the above problems, we propose the Tucker-O-Minus decomposition for multi-view subspace clustering. For employing the additional factors and correlations to construct a more compact and balanced tensor network, a simple way seems to replace the core factor of the Tucker structure with the TR format. 
However, as we mentioned above, the TR architecture suffers from a deficiency in exploring the correlations between the weakly-connected modes, which will result in the loss of essential clustering information in LRMVC. To this end, we propose the O-minus structure for LRMVC in the following three considerations. Firstly, from the perspective of two-point correlations in high-energy physics~\cite{billo2019two}, appending moderate lines with effective vertexes between two slightly correlated factors will strengthen their relationships. Similarly, we try to add the "bridge" with a tensor factor based on the TR structure to better capture the low rank information. Nonetheless, more links will generate more loops, which will cause huge difficulties in tensor contraction and the computational complexity burden of the tensor networks~\cite{cichocki2016low}. Simultaneously, since the number of samples $I \gg V$ in LRMVC, compared with higher-order information across different views, the information related to the correlations of instances requires more connections to be fully discovered. Accordingly, to efficiently explore the low rank information in LRMVC, the correlations related to samples (i.e., $I_{1}$, and $I_{3}$ in Fig.~\ref{fig: ndprocess}) are further strengthened by the special "bridge". This unique internal architecture is similar to "$\ominus$", thus namely O-Minus. And the whole architecture called Tucker-O-Minus decomposition is illustrated in Fig.~\ref{fig: ndprocess}. In this way, the low rank based multi-view clustering problem can be successfully solved with satisfying results. The main contributions of this paper are summarized as follows:
\begin{enumerate}
\item We propose the Tucker-O-Minus decomposition. Different from the existing tensor networks, it allows more valid interaction among nonadjacent factors and obtains a better low rank representation for high-dimensional data. Numerical experimental results on gray image reconstruction show that the performance of TOMD in exploring low rank information is superior to others. 
%\item We apply TOMD to a unified multi-view clustering framework, which can learn the representation and affinity matrix in a single step while preserving their correlations. Our proposed model, namely TOMD-MVC, utilizes the TOMD to capture the low-rank properties of self-representation tensor from multi-view data. The alternating direction method of multipliers (ADMM) is applied to solve the optimization model.
\item We apply TOMD to a unified multi-view clustering framework. The proposed model, namely TOMD-MVC, utilizes the TOMD to capture the low-rank properties of self-representation tensor from multi-view data. The alternating direction method of multipliers (ADMM) is applied to solve the optimization model.
\item We conduct extensive experiments on six real-world multi-view data sets to demonstrate the performance of our method. Compare with state-of-the-art models, TOMD-MVC achieves highly competitive or even better performance in terms of six evaluation metrics.
\end{enumerate}
The remainder of this paper is organized as follows. Section~\ref{sec:2} gives the used notations, mathematical backgrounds, and review the related works. The proposed tensor network is presented in details in Section~\ref{sec:4}. Section~\ref{sec:5} gives the multi-view clustering method based on the newly proposed tensor network. The experimental results are demonstrated and analyzed in Section~\ref{sec:6}. Section~\ref{sec:7} draws the conclusion finally.  

\section{Notations, Preliminaries, and Related Works}
\label{sec:2}
\subsection{Notations}
We give a brief introduction about notations and preliminaries in this section. Throughout this paper, we use lower case letters, bold lower case letters, bold upper case letters, and calligraphic letters to denote scalars, vectors, matrices and tensors, respectively, e.g., $a$, $\mathbf{a}$, $\mathbf{A}$ and $\mathcal{A}$. Some other frequently used notations are listed in Table~\ref{tab:notations}, where $\mathcal{A}$ is a 3rd-order tensor and $\mathbf{A}$ is a matrix.

\begin{table}[htbp]
\centering
\caption{Summary of notations in this paper.}
\scalebox{1.2}{
	\begin{tabular}{{l|l}}
		\hline
		Denotation         & Definition \\
		\hline
 		$\mathbf{A}_{i}$    		&$\mathbf{A}_{i}=\mathcal{A}(:,:,i)$\\                
		$a_{i_{1},i_{2},i_{3}}$     &The $(i_{1}, i_{2}, i_{3})$-th entry of tensor $\mathcal{A}$\\
		$\text{tr}(\mathbf{A})$	&$\text{tr}(\mathbf{A})=\sum_{i}a_{i,i}$	\\
		%$\|\mathbf{A}\|_{1}$           &$\|\mathbf{A}\|_{1}=\sum_{i_{1} i_{2}}\left|a_{i_{1},i_{2}}\right|$ \\    
		$\|\mathbf{A}\|_{\mathrm{F}}$  &$\|\mathbf{A}\|_{\mathrm{F}}=\sqrt{\sum_{i_{1} i_{2}} a_{i_{1},i_{2}}^{2}}$          \\
		%$\|\mathbf{A}\|_{*}$      	&Sum of singular values   \\   
		$\|\mathbf{A}\|_{\infty}$      &$\|\mathbf{A}\|_{\infty}=\max _{i_{1} i_{2}}\left|\mathbf{A}_{i_{1},i_{2}}\right|$ \\
		$\|\mathbf{A}\|_{2,1}$       	&$\|\mathbf{A}\|_{2,1}=\sum_{j}\|\mathbf{A}(:, j)\|_{2}$\\
		$\|\mathcal{A}\|_{\mathrm{F}}$     &$\|\mathcal{A}\|_{\mathrm{F}}=\sqrt{\sum_{i_{1} i_{2} i_{3}}\left|a_{i_{1},i_{2}, i_{3}}\right|^{2}}$   \\	
		\hline	
	\end{tabular}
}%\end{center}
\label{tab:notations}
\end{table}

\subsection{Preliminaries}

\begin{definition}
(\textbf{Mode-$n$ unfolding})~\cite{kolda2009tensor}
For tensor $\mathcal{A}\in\mathbb{R}^{I_1\times\cdots\times I_N}$, its matricization along the $n$-th mode is denoted as $\mathbf{A}_{(n)} \in \mathbb{R}^{I_{n} \times I_{1} I_{2} \cdots I_{n-1} I_{n+1} \cdots I_{N}}$. 
\end{definition}

\begin{definition}
(\textbf{$n$-unfolding})~\cite{cichocki2016low}
Given an $N$-way tensor $\mathcal{A}\in\mathbb{R}^{I_1\times\cdots\times I_N}$, the definition of its $n$-unfolding is expressed as $\mathbf{A}_{\langle n\rangle} \in \mathbb{R}^{I_{1} \cdots I_{n} \times I_{n+1} \cdots I_{N}}$. 
\end{definition}

\begin{definition}
(\textbf{Mode-$n$ product})~\cite{kolda2009tensor}
The mode-$n$ product of $\mathcal{A}\in \mathbb{R}^{I_{1} \times \cdots \times I_{N}}$ and matrix $\mathbf{B}\in\mathbb{R}^{J\times I_{n}}$ is defined as 
\begin{equation}
\mathcal{X}=\mathcal{A} \times_{n} \mathbf{B} \in \mathbb{R}^{I_{1} \times \cdots \times I_{n-1} \times J \times I_{n+1} \times \cdots \times I_{N}}.
\end{equation}
\end{definition}

\begin{definition}
(\textbf{Tensor Network Contraction})~\cite{ran2020tensor}
Given a tensor network composed of $N$ sub-tensors $\{\mathcal{G}^{(n)}\}(n=1,\cdots,N)$, then $\mathcal{G}$ is denoted as the tensor network contraction result of $\{\mathcal{G}^{(n)}\}(n=1,\cdots,N)$, whose general mathematical equation can be written as 
\begin{equation}
\begin{aligned}
\mathcal{G}=&\sum_{n=1}^{N}\sum_{r_{1}^{n},r_{2}^{n},\cdots}^{R_{1}^{n},R_{2}^{n},\cdots}\{\sum_{d_{1},d_{2}\cdots}^{D_{1},D_{2},\cdots} \prod_{n=1}^{N} \mathcal{G}^{(n)}(\{r_{1}^{n}, r_{2}^{n} \cdots, d_{1},d_{2} \cdots\})\}\\
=&\operatorname{TC}(\{\mathcal{G}^{(n)}\}_{n=1}^{N}),
\end{aligned}
\end{equation}
where $\{D_{1},D_{2},\cdots\}$ are geometrical indexes, each of which is shared normally two tensors and will be contracted, $\mathbb{R}=\{R_{1}^{1},R_{2}^{1},\cdots,R_{1}^{N},R_{2}^{N},\cdots\}$ are open bounds, each of which only belongs to one tensor. After contracting all the geometrical indexes, the $\mathcal{G}$ represents a $P$-th order tensor with $P$ the total number of the open indexes $\mathbb{R}$. 
\end{definition}

\subsection{Related Works}
\label{sec:3}
This section reviews some closely related works, including low-rank approximation based multi-view subspace clustering, and multi-view graph clustering.

\subsubsection{Low-rank Approximation Based Multi-view Subspace Clustering}
%\subsubsection{Low-rank Based Multi-view Subspace Clustering}
Subspace learning based multi-view clustering methods assume that each sample can be represented as a linear combination of all the samples \cite{sui2019sparse}.  

%Since the matrix based LRMVC ignore the high-order information hidden in multi-view data, Zhang et al. \cite{zhang2015low} consider all the subspace representations of the individual views as a high order structure, i.e. 3rd-order tensor, with unfolding form of tensor low-rank constraints. 
Chen et al. \cite{chen2021low} utilize the Tucker decomposition to obtain the "clean" representation tensor from view specific matrices. He et al. \cite{he2018self} introduce a tensor-based multi-linear multi-view clustering (MMC) method. Under the consideration of Canonical Polyadic Decomposition (CPD) \cite{kiers2000towards}, the MMC is able to explore the higher-order interaction information among the multiple views. Recently, tensor singular value decomposition (t-SVD) based tensor nuclear norm \cite{kilmer2008third,kilmer2011factorization} shows satisfying performance in capturing low rank information for 3rd-order tensor. Thus, many t-SVD based works \cite{xie2018unifying,wu2019essential,wu2020unified} are proposed for better capturing the high-order correlation hidden in 3rd-order tensor from multi-view data. To preserve the consensus across different views,  Xie et al. \cite{xie2018unifying} proposed the t-SVD-MSC model, which adopts the t-SVD based tensor nuclear norm to LRMVC. Consider that those self-representation based subspace clustering method will lead to high computation complexity, the essential tensor learning for multi-view spectral clustering (ETLMSC)~\cite{wu2019essential} with Markov chain based spectral clustering and the unified graph and low-rank tensor learning (UGLTL)~\cite{wu2020unified} have been proposed. Zheng et al. \cite{zheng2020constrained} develop a bi-linear factorization based clustering method with the orthogonality and low-rank constraint to effectively explore the consensus information of multi-view data. Work in \cite{xia2022tensorized} develops the tensorized bipartite graph learning for multi-view clustering (TBGL), which considers the inter-view and intra-view similarities with tensor Schatten $p$-norm and $\ell_{1,2}$-norm penalty, respectively. 
Yu et al. \cite{yu2020graph} propose NTR decomposition and the GNTR decomposition, which yield better performance than the state-of-the-art tensor based methods in clustering.

\subsubsection{Multi-view Graph Clustering}
Graph learning based multi-view clustering obtains a fusion graph across all views and uses the standard clustering algorithm to produce the final clusters \cite{yang2018multi}. Given a multi-view dataset, which has $V$ multiple views $\left\{\mathbf{X}_{v} \in \mathbb{R}^{C_{v} \times N}\right\}$ ($v=1,\cdots,V$), a general multi-view graph clustering optimization model can be formulated as follows: 
\begin{equation}
\begin{aligned}
&\min \limits_{\mathbf{M}_{i}} \sum \limits_{v=1}^{V} \sum \limits_{j=1}^{N} \frac{1}{2}\left\|(\mathbf{X}_{i})_{v}-(\mathbf{X}_{j})_{v}\right\|_2^{2} m_{i ,j}+\lambda \sum \limits_{j=1}^{N} m_{i, j}^{2} \\
&\text { s.~t.~} \mathbf{M}_{i}^{\mathrm{T}} \mathbf{1}=1, 0 \leq {m}_{i,j} \leq 1,
\end{aligned}
\label{eq:mgm}
\end{equation}
where $\mathbf{M}\in \mathbb{R}^{N \times N}$ is the affinity matrix, $\mathbf{M}_{i}$ is the  $i$-th column of matrix $\mathbf{M}$, $(\mathbf{X}_{i})_{v}$ denotes the $i$-th column of $\mathbf{X}_{v}$, and $\lambda$ is the trade-off parameter. 

By defining the graph Laplacian matrix $\mathbf{L}=\mathbf{D}-\left(\mathbf{M}+\mathbf{M}^{\mathrm{T}}\right)/2$, where $\mathbf{D}$ is a diagonal matrix whose $i$-th diagonal entry is $\sum_{j}\left(m_{i, j}+m_{j, i}\right) / 2$, (\ref{eq:mgm}) can be rewritten as
\begin{equation}
\begin{aligned}
&\min _{\mathbf{M}} \sum \limits_{v=1}^{V} \operatorname{tr}\left(\mathbf{X} \mathbf{L} \mathbf{X}^{\mathrm{T}}\right)+\lambda\|\mathbf{M}\|_{\text{F}}^{2} \\
&\text { s. t. } \mathbf{M}^{\mathrm{T}} \mathbf{1}=\mathbf{1}, \mathbf{0} \leq \mathbf{M} \leq \mathbf{1}.
\end{aligned}
\label{eq:trmgm}
\end{equation}

Based on~(\ref{eq:trmgm}), works in \cite{nie2016parameter, hou2017multi, nie2017self, zhuge2017unsupervised} use a parameter-free multiple graph clustering model to automatically learn the optimal weight for graphs. Moreover, Nie et al. introduce a multi-view learning model with adaptive neighbors \cite{nie2017multi}, which simultaneously performs clustering and local structure learning. Wang et al. introduce a general graph-based multi-view clustering (GMC) method \cite{wang2019gmc}, which does not only obtain the final clusters directly but also constructs the graph of each view and the fusion graph jointly, thus they can help each other mutually. 

\section{Tucker-O-Minus Decomposition}
\label{sec:4}
In this section, the mathematical formulation of Tucker-O-Minus decomposition with its corresponding solution is given in details.

\subsection{Decomposition Formulation}
Given a 4th-order tensor $\mathcal{X}\in\mathbb{R}^{I_1 \times I_2 \times I_3 \times I_4}$, the mathematical formulation of the TOMD is as follows:
\begin{equation}
\begin{aligned}
\mathcal{X}=&\{\sum_{r_{1},\cdots, r_{4}}^{R_{1},\cdots, R_{4}}(\sum_{d_{1},\cdots,d_{6}}^{D_{1},\cdots,D_{6}} \mathcal{G}^{(1)}(d_4,r_1,d_1,d_5) \mathcal{G}^{(2)}(d_1,r_2,d_2) \\
&\mathcal{G}^{(3)}(d_2,r_3,d_3,d_6) \mathcal{G}^{(4)}(d_3,r_4,d_4)\mathbf{G}^{(5)}(d_5,d_6))\}\\
&\times_{1} \mathbf{U}^{(1)} \times_{2}  \mathbf{U}^{(2)} \times_{3} \mathbf{U}^{(3)}  \times_{4} \mathbf{U}^{(4)}\\
=&\operatorname{TC}(\{\mathcal{G}^{(n)}\}_{n=1}^{5})\times_{1} \mathbf{U}^{(1)} \times_{2}  \mathbf{U}^{(2)} \times_{3} \mathbf{U}^{(3)}  \times_{4} \mathbf{U}^{(4)},
\end{aligned}
\label{eq:dtnmodel}
\end{equation}
where $\{D_{1},\cdots,D_{6}\}$ are geometrical indexes, each of which is shared two tensors and should be contracted. $\mathbf{U}^{(i)}\in\mathbb{R}^{I_{i}\times R_{i}}(i=1,\cdots,4)$ are factor matrices, $\mathcal{G}^{(n)}$, $n=1,\cdots,5$, are the factor tensors. Specifically, $\mathcal{G}^{(1)}\in\mathbb{R}^{D_4 \times R_1 \times D_1\times D_5}$, $\mathcal{G}^{(2)}\in\mathbb{R}^{D_1 \times R_2 \times D_2}$, $\mathcal{G}^{(3)}\in\mathbb{R}^{D_2 \times R_3 \times D_3\times D_6 }$, $\mathcal{G}^{(4)}\in\mathbb{R}^{D_3 \times R_4 \times D_4}$, and $\mathbf{G}^{(5)}\in\mathbb{R}^{D_5 \times D_6}$. $[R_{1},\cdots,R_{4},D_{1},\cdots,D_{6}]^\text{T}$ is denoted as the rank of TOMD. 
 
From the graphical illustration of TOMD in Fig.~\ref{fig: ndprocess}, it can be clearly seen that the overall architecture of our proposed tensor network is relatively compact and balanced. 
In this way, based on the Tucker format, O-minus structure is further applied to enhance the relationships among the adjacent and non-adjacent modes of $\mathcal{X}$, simultaneously. In addition, the computational complexity of tensor network contraction can be controlled. 
%For Tucker decomposition, an efficient and simple scheme to further reduce the size of the core tensor is employing the distributed tensor networks~\cite{cichocki2016low}. In this way, we adopt the $\ominus$-like design to control the tensor ranks. Besides, another matrix $\mathbf{G}^{(5)}$ is used to bridge two nonadjacent factors to strengthen the interactions, which can help to avoid the rapid increase of factors' ranks. 

\subsection{ALS algorithm}
To compute the Tucker-O-Minus decomposition, we employ the alternation least squares (ALS) algorithm, which has been widely used for CPD, Tucker decomposition and TR decomposition~\cite{carroll1970analysis, kroonenberg1980principal, zhao2016tensor}. Supposing that we have a 4th-order tensor $\mathcal{X} \in \mathbb{R}^{I_1 \times I_2 \times I_3 \times I_4}$, the optimization model for approximation a tensor $ \mathcal{X} $ by the proposed TOMD can be formulated as follows:
\begin{equation}
%\min \limits_{\{{\mathcal{G}}^{(n)}\}_{n=1}^{5},\atop\{{\mathbf{U}}^{(n)}\}_{n=1}^{4}}\|\mathcal{X}-\operatorname{TC}(\{\mathcal{G}^{(n)}\}_{n=1}^{5})\times_{1} \mathbf{U}^{(1)} \times_{2}  \cdots  \times_{4} \mathbf{U}^{(4)}\|_{\mathrm{F}}.
\min \limits_{\substack{\{{\mathcal{G}}^{(n)}\}_{n=1}^{5}\\\{{\mathbf{U}}^{(n)}\}_{n=1}^{4}}}\|\mathcal{X}-\operatorname{TC}(\{\mathcal{G}^{(n)}\}_{n=1}^{5})\times_{1} \mathbf{U}^{(1)} \times_{2}  \cdots  \times_{4} \mathbf{U}^{(4)}\|_{\mathrm{F}}.
\label{eq:alsdtn}
\end{equation}

\begin{theorem}
\label{th:1}
Given a Tucker-O-Minus decomposition $\mathcal{X}=\operatorname{TC}(\{\mathcal{G}^{(n)}\}_{n=1}^{5})\times_{1} \mathbf{U}^{(1)} \times_{2}  \mathbf{U}^{(2)} \times_{3} \mathbf{U}^{(3)}  \times_{4} \mathbf{U}^{(4)}$, the corresponding mode-$n$ unfolding based matrix equivalence of (\ref{eq:dtnmodel}) can be written as $\mathbf{X}_{(n)}=\mathbf{U}^{(n)}(\mathbf{A}_{(n)})$, or $\mathbf{X}_{(n)}=\mathbf{U}^{(n)}(\mathbf{G}_{(2)}^{(n)})(\mathbf{A}^{\neq n})$,
% \begin{equation}
% \mathbf{X}_{(n)}=\mathbf{U}^{(n)}(\mathbf{A}_{(n)}),
% \end{equation}
% or 
% \begin{equation}
% \mathbf{X}_{(n)}=\mathbf{U}^{(n)}(\mathbf{G}_{(2)}^{(n)})(\mathbf{A}^{\neq n}),
% \label{eq:the12}
% \end{equation}
where $\mathbf{A}^{\neq n}=\mathbf{A}_{\langle 3\rangle}^{\neq n}~(n=1~\text{or}~3)$, and $\mathbf{A}^{\neq n}=\mathbf{A}_{\langle 2\rangle}^{\neq n}~(n=2~\text{or}~4)$. Moreover, we have $\mathcal{A}=\mathcal{G}\times_{1} \mathbf{U}^{(1)}  \cdots \times_{n-1} \mathbf{U}^{(n-1)} \times_{n+1} \mathbf{U}^{(n+1)}  \cdots \times_{4} \mathbf{U}^{(4)}$, and $\mathcal{A}=\mathcal{G}\times_{1} \mathbf{U}^{(1)}  \cdots \times_{n-1} \mathbf{U}^{(n-1)} \times_{n+1} \mathbf{U}^{(n+1)}  \cdots \times_{4} \mathbf{U}^{(4)}$,
%$\mathcal{A}$ and $\mathcal{A}^{\neq n}$ as follows: 
% \begin{equation*}
%  \mathcal{A}=\mathcal{G}\times_{1} \mathbf{U}^{(1)}  \cdots \times_{n-1} \mathbf{U}^{(n-1)} \times_{n+1} \mathbf{U}^{(n+1)}  \cdots \times_{4} \mathbf{U}^{(4)},
%  \end{equation*}
%  and
%  \begin{equation*}
%  \mathcal{A}^{\neq n}=\mathcal{G}^{\neq n}\times_{1} \mathbf{U}^{(1)} \cdots \times_{n-1} \mathbf{U}^{(n-1)} \times_{n+1} \mathbf{U}^{(n+1)} \cdots \times_{4} \mathbf{U}^{(4)},
%  \end{equation*}
respectively, where $\mathcal{G}=\operatorname{TC}(\{\mathcal{G}^{(n)}\}_{n=1}^{5})$ and $\mathcal{G}^{\neq n}=\operatorname{TC}(\{\mathcal{G}^{(n_{1})}\}_{n_{1}\neq n}^{5})$. $\mathbf{A}_{\langle 2\rangle}^{\neq n}$ and $\mathbf{A}_{\langle 3\rangle}^{\neq n}$ are the $n$-unfolding matrices of $ \mathcal{A}^{\neq n}$.
\end{theorem}

\begin{theorem}
\label{th:2}
Suppose that we have a Tucker-O-Minus decomposition $\mathcal{X}=\operatorname{TC}(\{\mathcal{G}^{(n)}\}_{n=1}^{5})\times_{1} \mathbf{U}^{(1)} \times_{2}  \mathbf{U}^{(2)} \times_{3} \mathbf{U}^{(3)}  \times_{4} \mathbf{U}^{(4)}$, and its vectorized form can be formulated as $\mathbf{x}=\mathbf{g}^{(5)}\tilde{\mathbf{A}}_{\langle 2\rangle}$, 
% \begin{equation}
% \mathbf{x}=\mathbf{g}^{(5)}\tilde{\mathbf{A}}_{\langle 2\rangle},
% \end{equation}
where $\mathbf{x}\in \mathbb{R}^{1\times I_{1}I_{2}I_{3}I_{4}}$, $\mathbf{g}^{(5)}\in \mathbb{R}^{1\times D_{5 }D_{6}}$ are the vectorized form of $\mathcal{X}$ and $\mathbf{G}^{(5)}$, respectively. Furthermore, $\tilde{\mathcal{A}}$ can be calculated by $\tilde{\mathcal{A}}=\mathcal{G}^{\neq 5}\times_{3} \mathbf{U}^{(1)} \times_{4}  \cdots  \times_{6} \mathbf{U}^{(4)}\in\mathbb{R}^{D_{5} \times D_{6} \times I_{1} \times \cdots \times I_{4}}$, 
% \begin{equation*}
% \tilde{\mathcal{A}}=\mathcal{G}^{\neq 5}\times_{3} \mathbf{U}^{(1)} \times_{4}  \cdots  \times_{6} \mathbf{U}^{(4)}\in\mathbb{R}^{D_{5} \times D_{6} \times I_{1} \times \cdots \times I_{4}},
%  \end{equation*}
where $\mathcal{G}^{\neq 5}=\operatorname{TC}(\{\mathcal{G}^{(n)}\}_{n=1}^{4}) \in \mathbb{R}^{D_{5} \times D_{6} \times R_{1} \times R_{2} \times R_{3} \times R_{4}}$. $\tilde{\mathbf{A}}_{\langle 2\rangle}$ is the $n$-unfolding of $\tilde{\mathcal{A}}$.
 \end{theorem}

%From Fig.~\ref{fig: ndprocess}, Theorem~\ref{th:1}, the optimization problem (\ref{eq:alsdtn}) can be changed into a few subproblems by alternating least squares method. In detail, fixing all but one factor, the problem can reduce to three least-squares problems:
From Fig.~\ref{fig: ndprocess}, Theorem~\ref{th:1} and Theorem~\ref{th:2}, the optimization problem (\ref{eq:alsdtn}) can be changed into a few subproblems by alternating least squares method. In detail, fixing all but one factor, the problem can reduce to three least-squares problems:
\begin{enumerate}%[\IEEEsetlabelwidth{12)}]
\item Subproblem with respect to $\mathbf{U}^{(n)}(n=1,\cdots,4)$:
\begin{equation}
\min \limits_{\mathbf{U}^{(n)}}\|\mathbf{X}_{(n)}-\mathbf{U}^{(n)}(\mathbf{A}_{(n)})\|_{\text{F}}.
\end{equation}

\item Subproblem with respect to $\mathcal{G}^{(n)}(n=1,\cdots,4)$:
\begin{equation}
\min \limits_{\mathbf{G}_{(2)}^{(n)}}\|\mathbf{X}_{(n)}-\mathbf{U}^{(n)}(\mathbf{G}_{(2)}^{(n)})(\mathbf{A}^{\neq n})\|_{\text{F}}.
\end{equation}

\item Subproblem with respect to $\mathcal{G}^{(5)}$:
\begin{equation}
\min \limits_{\mathbf{g}^{(5)}}\|\mathbf{x}-(\mathbf{g}^{(5)})\tilde{\mathbf{A}}_{\langle 2\rangle}\|_{\text{F}}.
\end{equation}
% 12 items total
\end{enumerate}

Moreover, 9 factors (i.e., $\{\mathcal{G}^{(n)}\}_{n=1}^{5},\{{\mathbf{U}}^{(n)}\}_{n=1}^{4}$) can be initialized using the randomization method or truncated SVD with specific values of ranks. In our experiments, we choose the latter way to initialize all factors, and the stopping condition is $\|\mathcal{X}_{\text{last}}-\mathcal{X}_{\text{new}}\|_{\text{F}}/\|\mathcal{X}_{\text{last}}\|_{\text{F}} \leq \text{tol}_{\text {als}},$
% defined as follows:
% \begin{equation}
% \frac{\left\|\mathcal{X}_{\text{last}}-\mathcal{X}_{\text{new}}\right\|_{\text{F}}}{\left\|\mathcal{X}_{\text{last}}\right\|_{\text{F}}} \leq \text{tol}_{\text {als}},
% \label{eq:stopcon}
% \end{equation}
where $\mathcal{X}_{\text{last}}$ is the value at the last iteration, $\mathcal{X}_{\text{new}}$ is constructed from the new iteration, and $\text{tol}_{\text {als}}$ is a predefined threshold.

By denoting $\Phi({\cdot,\mathbf{s}})$ as an operator that reshapes a tensor using the size vector $\mathbf{s}$, the overall procedure of TOMD-ALS algorithm can be found in Algorithm~\ref{algo:1}.
%By denoting $\Phi({\cdot,\mathbf{s}})$ as an operator that reshapes a tensor using the size vector $\mathbf{s}$, the overall procedure of TOMD-ALS algorithm\footnote{To make it better understanding, the iteration number $i$ is omitted in the process of algorithm} can be found in Algorithm~\ref{algo:1}.

\begin{center}
		\begin{algorithm}[tb]
			\caption{TOMD-ALS}
			\begin{algorithmic}
				\STATE \textbf{Input}: A 4th-order data tensor $\mathcal{X} \in \mathbb{R}^{I_1 \times I_2 \times I_3 \times I_4}$, predefined TOMD rank, maximum number of iterations ($\text{iter}_{\text{max}}$), and the threshold for stopping the algorithm $\text{tol}_{\text{als}}=10^{-12}$, $f_{1}=0$, $i=0$.
				\STATE \textbf{Initialization}: Initialize $\mathcal{G}^{(1)},\cdots,\mathcal{G}^{(4)},\mathbf{G}^{(5)},\mathbf{U}^{(1)},\cdots,\mathbf{U}^{(4)}$.
				\STATE $\mathcal{X}_{\text{new}}=\mathcal{G}\times_{1} \mathbf{U}^{(1)} \times_{2} \cdots  \times_{4} \mathbf{U}^{(4)}$;
				\WHILE{$i \leq \text{iter}_{\max }$}
				\STATE $\mathcal{X}_{\text{last}}=\mathcal{X}_{\text{new}}$;
				\FOR{ $n=1:4$}
				\STATE $\mathcal{A} =\mathcal{G}\times_{1} \mathbf{U}^{(1)} \cdots \times_{n-1}  \mathbf{U}^{(n-1)} \times_{n+1} \mathbf{U}^{(n+1)}\cdots \times_{4} \mathbf{U}^{(4)}$;
				\STATE $\mathbf{U}^{(n)}\leftarrow \arg \min \|\mathbf{X}_{(n)}-\mathbf{U}^{(n)}(\mathbf{A}_{(n)})\|_{\mathrm{F}}$;
				\ENDFOR
                			\FOR{$n=1:4$}
				\STATE $\mathcal{A}^{\neq n} =\mathcal{G}^{\neq n}\times_{1} \mathbf{U}^{(1)} \times_{2} \cdots \times_{n-1} \mathbf{U}^{(n-1)}\times_{n+1} \mathbf{U}^{(n+1)}\times_{n+2} \cdots \times_{4} \mathbf{U}^{(4)}$;
				\STATE $\mathbf{G}_{(2)}^{(n)}\leftarrow \arg \min \|\mathbf{X}_{(n)}-\mathbf{U}^{(n)}(\mathbf{G}_{(2)}^{(n)})(\mathbf{A}^{\neq n})\|_{\mathrm{F}}$;
				\STATE $\mathcal{G}^{(n)}$=$\Phi(\mathbf{G}_{(2)}^{(n)}$, $\operatorname{size}(\mathcal{G}^{(n)}$));
				\ENDFOR
				\STATE $\tilde{\mathcal{A}}= \mathcal{G}^{\neq 5}\times_{1} \mathbf{U}^{(1)} \times_{2} \cdots  \times_{4} \mathbf{U}^{(4)}$;
				\STATE $\tilde{\mathbf{A}}$=$\Phi(\tilde{\mathcal{A}},[\sim,I_1 \cdots I_4]$), $\mathbf{x}$=$\Phi(\mathcal{X},[\sim,I_1 \cdots I_4])$;
				\STATE $\mathbf{g}^{(5)}\leftarrow \argmin \|\mathbf{x}-(\mathbf{g}^{(5)})\tilde{\mathbf{A}}_{\langle 2\rangle}\|_{\mathrm{F}}$;
				\STATE $\mathbf{G}^{(5)}$=$\Phi(\mathbf{g}^{(5)}$, $\operatorname{size}(\mathcal{G}^{(5)}$));
				\STATE $\mathcal{X}_{\text{new}}=\mathcal{G}\times_{1} \mathbf{U}^{(1)} \times_{2} \cdots  \times_{4} \mathbf{U}^{(4)}$;
				\STATE $f_{1}=\frac{\left\|\mathcal{X}_{\text{last}}-\mathcal{X}_{\text{new}}\right\|_{\mathrm{F}}}{\left\|\mathcal{X}_{\text{last}}\right\|_{\mathrm{F}}}$;
				\IF{$f_{1}  \leq \text{tol}_{\text {als }}$}
				\STATE break
				\ENDIF
				\STATE $i=i+1$;
				\ENDWHILE		
				\STATE \textbf{Output:} $\mathcal{G}^{(1)}, \cdots,\mathcal{G}^{(4)},\mathbf{G}^{(5)},\mathbf{U}^{(1)}, \cdots, \mathbf{U}^{(4)}$.
			\end{algorithmic}
			\label{algo:1}
		\end{algorithm}
	\end{center}

\section{Tucker-O-Minus decomposition for low rank multi-view clustering model}
\label{sec:5}
In this section, we firstly apply TOMD into the low rank tensor optimization model for multi-view clustering. The alternating direction method of multipliers is used to solve it. And its computational complexity is briefly analyzed.

\subsection{Optimization Model}
To better exploit the low rank data structure in multi-view clustering, we apply the proposed TOMD to the optimization model for clustering. Given a dataset that has $V$ views $\left\{\mathbf{X}_{v} \in \mathbb{R}^{C_{v} \times N}\right\}$ $, v = 1, \cdots, V$, where $C_{v}$ is the feature dimension of the $v$-th feature and $N$ is the number of data samples in each view. The proposed optimization model based on TOMD for multi-view cluster (TOMD-MVC) can be formulated as follows:
\begin{equation}\label{eq:1}
\begin{aligned}
% &\min \limits_{\mbox{\tiny$\begin{array}{c}
%   \mathcal{Z}, \mathbf{E}, \mathbf{M}\\
%   \mathcal{G}^{(1)},\cdots, \mathcal{G}^{(4)}, \mathbf{G}^{(5)}\\
%   \mathbf{U}^{(1)}, \cdots, \mathbf{U}^{(4)}\end{array}$}}\mu \sum_{v=1}^{V} \operatorname{tr}\left(\mathbf{Z}_{v}^{\mathrm{T}} \mathbf{L} \mathbf{Z}_{v}\right)+\lambda\|\mathbf{M}\|_{\mathrm{F}}^{2}+ \| \mathbf{E}\|_{2,1} \\
% &~\text {s.~t.}~\mathbf{X}_{v}=\mathbf{X}_{v} \mathbf{Z}_{v}+\mathbf{E}_{v},~v=1,2, \cdots, V, \\
% &~~~~\mathbf{M}^{\mathrm{T}} \mathbf{1}=\mathbf{1}, \mathbf{0} \leq  \mathbf{M} \leq \mathbf{1},\\
% &~~~~\mathcal{Z}=\Phi(\operatorname{TC}(\{\mathcal{G}^{(n)}\}_{n=1}^{5})\times_{1} \mathbf{U}^{(1)}   \cdots \times_{4} \mathbf{U}^{(4)},[N,N,V]), \\
% &~~~~\mathcal{Z}=\operatorname{\Omega}\left(\mathbf{Z}_{1},\mathbf{Z}_{2}, \cdots, \mathbf{Z}_{V}\right),\mathbf{E}=\left[ \mathbf{E}_{1} ; \mathbf{E}_{2} ; \cdots ;  \mathbf{E}_{V}\right],
&\min \limits_{\substack{\mathcal{Z}, \mathbf{E}, \mathbf{M}\\
  \{\mathcal{G}^{(n)}\}_{n=1}^{5},\{\mathbf{U}^{(n)}\}_{n=1}^{4}}}\mu \sum_{v=1}^{V} \operatorname{tr}\left(\mathbf{Z}_{v}^{\mathrm{T}} \mathbf{L} \mathbf{Z}_{v}\right)+\lambda\|\mathbf{M}\|_{\mathrm{F}}^{2}+ \| \mathbf{E}\|_{2,1} \\
&~\text {s.~t.}~\mathbf{X}_{v}=\mathbf{X}_{v} \mathbf{Z}_{v}+\mathbf{E}_{v},~v=1,2, \cdots, V, \\
&~~~~\mathbf{M}^{\mathrm{T}} \mathbf{1}=\mathbf{1}, \mathbf{0} \leq  \mathbf{M} \leq \mathbf{1},\\
&~~~~\mathcal{Z}=\Phi(\operatorname{TC}(\{\mathcal{G}^{(n)}\}_{n=1}^{5})\times_{1} \mathbf{U}^{(1)}   \cdots \times_{4} \mathbf{U}^{(4)},[N,N,V]), \\
&~~~~\mathcal{Z}=\operatorname{\Omega}\left(\mathbf{Z}_{1},\mathbf{Z}_{2}, \cdots, \mathbf{Z}_{V}\right),\mathbf{E}=\left[ \mathbf{E}_{1} ; \mathbf{E}_{2} ; \cdots ;  \mathbf{E}_{V}\right],
\end{aligned}
\end{equation}
%where $\mathcal{G}^{(1)}\in\mathbb{R}^{D_4 \times R_1 \times D_1\times D_5}$, $\mathcal{G}^{(2)}\in\mathbb{R}^{D_1 \times R_2 \times D_2}$, $\mathcal{G}^{(3)}\in\mathbb{R}^{D_2 \times R_3 \times D_3\times D_6 }$, $\mathcal{G}^{(4)}\in\mathbb{R}^{D_3 \times R_4 \times D_4}$, and $\mathbf{G}^{(5)}\in\mathbb{R}^{D_5 \times D_6}$. $\{D_{1},\cdots,D_{6}\}$ are the factors, $[R_{1},\cdots,R_{4},D_{1},\cdots,D_{6}]^\text{T}$ denotes the rank of $\mathcal{Z}$, and $\mathbf{L}$ is the graph Laplacian matrix of $\mathbf{M}$. $\mathbf{E}=\left[ \mathbf{E}_{1} ;  \mathbf{E}_{2} ; \cdots ;  \mathbf{E}_{V}\right]$ makes all the error matrices $\{\mathbf{E}_{v}\}$ $(v=1,\cdots,V)$ vertically connected along the columns, and function $\operatorname{\Omega}(\cdot)$ constructs the matrices $\{\mathbf{Z}_{v}\}(v=1,\cdots,V)$ into a 3rd-order tensor. Apart from that, $\mu$ and $\lambda$ are positive penalty scalars. 
where $\{\mathcal{G}^{(n)}\}_{n=1}^{5}$ are tensor factors, and $\mathbf{L}$ is the graph Laplacian matrix of $\mathbf{M}$. $\mathbf{E}=\left[ \mathbf{E}_{1} ;  \mathbf{E}_{2} ; \cdots ;  \mathbf{E}_{V}\right]$ makes all the error matrices $\{\mathbf{E}_{v}\}$ $(v=1,\cdots,V)$ vertically connected along the columns, and function $\operatorname{\Omega}(\cdot)$ constructs the matrices $\{\mathbf{Z}_{v}\}(v=1,\cdots,V)$ into a 3rd-order tensor. Apart from that, $\mu$ and $\lambda$ are positive penalty scalars. 

After obtaining the affinity matrix $\mathbf{M}$, K-means~\cite{hartigan1979algorithm}, spectral clustering~\cite{ng2002spectral} and other clustering methods can be utilized to yield the final clustering results.

\subsection{Solution}
The optimization problem~(\ref{eq:1}) can be solved by ADMM. Since variable $\mathcal{Z}$ has appeared in the objective function and three constraints simultaneously, we introduce an extra variable $\mathcal{S}$ to make $\mathcal{Z}$ separable, which result in the equivalence of (\ref{eq:1}) as follows:
\begin{equation}\label{eq:2}
\begin{aligned}
% &\min \limits_{\mbox{\tiny$\begin{array}{c}
%   \mathcal{Z}, \mathcal{S},\mathbf{E}, \mathbf{M}\\
%   \mathcal{G}^{(1)},\cdots, \mathcal{G}^{(4)}, \mathbf{G}^{(5)}\\
% \mathbf{U}^{(1)}, \cdots, \mathbf{U}^{(4)}\end{array}$}}\mu \sum_{v=1}^{V} \operatorname{tr}(\mathbf{Z}_{v}^{\mathrm{T}} \mathbf{L} \mathbf{Z}_{v})+\lambda\|\mathbf{M}\|_{\mathrm{F}}^{2}+ \| \mathbf{E}\|_{2,1} \\
% &~~~~\text{ s.~t.} ~\mathbf{X}_{v}=\mathbf{X}_{v} \mathbf{S}_{v}+\mathbf{E}_{v},~v=1,2, \cdots, V, \\
% &~~~~~\mathbf{M}^{\mathrm{T}} \mathbf{1}=\mathbf{1}, \mathbf{0} \leq  \mathbf{M} \leq \mathbf{1},\\
% &~~~~~\mathcal{Z}=\Phi(\operatorname{TC}(\{\mathcal{G}^{(n)}\}_{n=1}^{5})\times_{1} \mathbf{U}^{(1)}   \cdots \times_{4} \mathbf{U}^{(4)},[N,N,V]),\\
% &~~~~~\mathcal{Z}=\mathcal{S},\\
% &~~~~~\mathcal{Z}=\operatorname{\Omega}\left(\mathbf{Z}_{1}, \mathbf{Z}_{2}, \cdots, \mathbf{Z}_{V}\right),\mathbf{E}=\left[\mathbf{E}_{1} ; \mathbf{E}_{2} ; \cdots ; \mathbf{E}_{V}\right].
&\min \limits_{\substack{\mathcal{Z}, \mathbf{E}, \mathbf{M}\\
  \{\mathcal{G}^{(n)}\}_{n=1}^{5},\{\mathbf{U}^{(n)}\}_{n=1}^{4}}}\mu \sum_{v=1}^{V} \operatorname{tr}(\mathbf{Z}_{v}^{\mathrm{T}} \mathbf{L} \mathbf{Z}_{v})+\lambda\|\mathbf{M}\|_{\mathrm{F}}^{2}+ \| \mathbf{E}\|_{2,1} \\
&~~~~\text{ s.~t.} ~\mathbf{X}_{v}=\mathbf{X}_{v} \mathbf{S}_{v}+\mathbf{E}_{v},~v=1,2, \cdots, V, \\
&~~~~~\mathbf{M}^{\mathrm{T}} \mathbf{1}=\mathbf{1}, \mathbf{0} \leq  \mathbf{M} \leq \mathbf{1},\\
&~~~~~\mathcal{Z}=\Phi(\operatorname{TC}(\{\mathcal{G}^{(n)}\}_{n=1}^{5})\times_{1} \mathbf{U}^{(1)}   \cdots \times_{4} \mathbf{U}^{(4)},[N,N,V]),\\
&~~~~~\mathcal{Z}=\mathcal{S},\\
&~~~~~\mathcal{Z}=\operatorname{\Omega}\left(\mathbf{Z}_{1}, \mathbf{Z}_{2}, \cdots, \mathbf{Z}_{V}\right),\mathbf{E}=\left[\mathbf{E}_{1} ; \mathbf{E}_{2} ; \cdots ; \mathbf{E}_{V}\right].
\end{aligned}
\end{equation}
The augmented Lagrangian function of optimization model (\ref{eq:2}) is

% \begin{equation}\label{eq:3}
% \begin{aligned}
% \operatorname{L}(\mathcal{Z}, \mathcal{S}, \mathbf{E}, \mathbf{M}; \mathcal{W}, \mathcal{Y})&=\lambda\|\mathbf{M}\|_{\mathrm{F}}^{2}+\|\mathbf{E}\|_{2,1}\\
% &+\sum_{v=1}^{V}(\mu \operatorname{tr}(\mathbf{S}_{v}^{\mathrm{T}} \mathbf{L} \mathbf{S}_{v}) \\
% &+\langle\mathbf{W}_{v}, \mathbf{X}_{v}-\mathbf{X}_{v} \mathbf{S}_{v}-\mathbf{E}_{v}\rangle\\
% &+\frac{\tau}{2}\|\mathbf{X}_{v}-\mathbf{X}_{v} \mathbf{S}_{v}-\mathbf{E}_{v}\|_{\mathrm{F}}^{2})\\
% &+\langle \mathcal{Y}, \mathcal{Z}-\mathcal{S}\rangle+\frac{\tau}{2}\|\mathcal{Z}-\mathcal{S}\|_{\mathrm{F}}^{2},
% \end{aligned}
% \end{equation}
\begin{equation}\label{eq:3}
\begin{aligned}
&\operatorname{L}(\mathcal{Z}, \mathcal{S}, \mathbf{E}, \mathbf{M}; \mathcal{W}, \mathcal{Y})=\lambda\|\mathbf{M}\|_{\mathrm{F}}^{2}+\|\mathbf{E}\|_{2,1}+\sum_{v=1}^{V}(\mu \operatorname{tr}(\mathbf{S}_{v}^{\mathrm{T}} \mathbf{L} \mathbf{S}_{v}) \\
&+\langle\mathbf{W}_{v}, \mathbf{X}_{v}-\mathbf{X}_{v} \mathbf{S}_{v}-\mathbf{E}_{v}\rangle+\frac{\tau}{2}\|\mathbf{X}_{v}-\mathbf{X}_{v} \mathbf{S}_{v}-\mathbf{E}_{v}\|_{\mathrm{F}}^{2})\\
&+\langle \mathcal{Y}, \mathcal{Z}-\mathcal{S}\rangle+\frac{\tau}{2}\|\mathcal{Z}-\mathcal{S}\|_{\mathrm{F}}^{2},
\end{aligned}
\end{equation}
where $\mathcal{W}$ and  $\mathcal{Y}$ are Lagrangian multipliers corresponding to two equations, $\langle\cdot, \cdot\rangle$ denotes the inner product, and $\tau$ is the penalty parameter. ADMM alternately updates each variable as follows.

\textbf{Solving $\mathcal{Z}$-subproblem:} 
With the other variables fixed but let $\mathcal{Z}$ be the optimization variable, (\ref{eq:2}) becomes 
\begin{equation}
\begin{aligned}
% &\min \limits_{\mbox{\tiny$\begin{array}{c}
%   \mathcal{Z}\\
%   \mathcal{G}^{(1)},\cdots, \mathcal{G}^{(4)},\mathbf{G}^{(5)}\\
%  \mathbf{U}^{(1)}, \cdots, \mathbf{U}^{(4)}\end{array}$}}\|\mathcal{Z}-(\mathcal{S}^{t}-\frac{\mathcal{Y}^{t}}{\tau^{t}})\|_{\mathrm{F}}^{2} \\
% &\text {s. t.}~\mathcal{Z}=\Phi(\operatorname{TC}(\{\mathcal{G}^{(n)}\}_{n=1}^{5})\times_{1} \mathbf{U}^{(1)} \cdots \times_{4} \mathbf{U}^{(4)},[N,N,V]), 
&\min \limits_{\substack{\mathcal{Z},\{\mathcal{G}^{(n)}\}_{n=1}^{5},\{\mathbf{U}^{(n)}\}_{n=1}^{4}}}\|\mathcal{Z}-(\mathcal{S}^{t}-\frac{\mathcal{Y}^{t}}{\tau^{t}})\|_{\mathrm{F}}^{2} \\
&\text {s. t.}~\mathcal{Z}=\Phi(\operatorname{TC}(\{\mathcal{G}^{(n)}\}_{n=1}^{5})\times_{1} \mathbf{U}^{(1)} \cdots \times_{4} \mathbf{U}^{(4)},[N,N,V]), 
\end{aligned}
\end{equation}
where $t$ is the number of iteration. This problem can be solved by TOMD-ALS to obtain the factors $\tilde{\mathcal{G}}^{(i)}(i=1,\cdots,5)$ 
%$(\tilde{\mathcal{G}}^{(1)}, \cdots,\tilde{\mathcal{G}}^{(4)},\tilde{\mathbf{G}}^{(5)})$ 
and $\tilde{\mathbf{U}}^{(i)}(i=1,\cdots, 4)$ of tensor $(\mathcal{S}^{t}-\frac{\mathcal{Y}^{t}}{\tau^{t}})$. Accordingly, the $\mathcal{Z}$-subproblem has following solution:  
\begin{equation}
\mathcal{Z}^{t+1}=\Phi(\operatorname{TC}(\{\tilde{\mathcal{G}^{(n)}}\}_{n=1}^{5})\times_{1} \tilde{\mathbf{U}}^{(1)} \times_{2} \cdots \times_{4} \tilde{\mathbf{U}}^{(4)},[N,N,V]).
\label{eq:upzpro}
\end{equation}

\textbf{Solving $\mathcal{S}$-subproblem:} 
Similarly, letting $\mathcal{S}$ be the only variable, (\ref{eq:2}) becomes 
\begin{equation}
\begin{aligned}
\mathcal{S}^{t+1}&=\underset{\mathcal{S}}{\operatorname{argmin}} \sum \limits_{v=1}^{V}(\frac{\tau^{t}}{2}\|\mathbf{X}_{v}-\mathbf{X}_{v} \mathbf{S}_{v}-\mathbf{E}_{v}^{t}+\frac{\mathbf{W}_{v}^{t}}{\tau^{t}}\|_{\mathrm{F}}^{2} \\
&+\mu \operatorname{tr}(\mathbf{S}_{v}^{\mathrm{T}} \mathbf{L} \mathbf{S}_{v}))+\frac{\tau^{t}}{2}\|\mathcal{Z}^{t+1}-\mathcal{S}+\frac{\mathcal{Y}^{t}}{\tau^{t}}\|_{\mathrm{F}}^{2}.
\end{aligned}
\end{equation}

By separating above problem into $V$ subproblems, each of which can be solved by:
\begin{equation}
\begin{split}
\mathbf{S}_{v}^{t+1}&=(\tau^{t}(\mathbf{t}+\mathbf{X}_{v}^{\mathrm{T}} \mathbf{X}_{v})+2 \mu \mathbf{L})^{-1}(\tau^{t} \mathbf{Z}_{v}^{t+1} \\
&+\mathbf{Y}_{v}^{t}+\tau^{t} \mathbf{X}_{v}^{\mathrm{T}}(\mathbf{X}_{v}-\mathbf{E}_{v}^{t}+\frac{\mathbf{W}_{v}^{t}}{\tau^{t}})).
\end{split}
\end{equation}

\textbf{Solving $\mathbf{E}$-subproblem:} 
The optimization subproblem with respect to $\mathbf{E}$ is:
\begin{equation}\label{eq:upeem}
\begin{aligned}
\mathbf{E}^{t+1}&=\underset{\mathbf{E}}{\operatorname{argmin}} \frac{1}{\tau^{t}}\|\mathbf{E}\|_{2,1}\\
&+\frac{1}{2} \sum_{v=1}^{V}\|\mathbf{E}_{v}-(\mathbf{X}_{v}-\mathbf{X}_{v} \mathbf{S}_{v}^{t+1}+\frac{\mathbf{W}_{v}^{t}}{\tau^{t}})\|_{\mathrm{F}}^{2}.
\end{aligned}
\end{equation}

Let $\mathbf{H}_{v}^{t}=\mathbf{X}_{v}-\mathbf{X}_{v} \mathbf{S}_{v}^{t+1}+\frac{\mathbf{W}_{v}^{t}}{\tau^{t}}$, then $\mathbf{H}^{t}$ is constructed by vertically concatenating the matrices $\left\{\mathbf{H}_{v}\right\}, v = 1,\cdots,V$. As it is suggested in~\cite{liu2012robust}, (\ref{eq:upeem}) has a closed-form solution:
%\begin{equation}
%\mathbf{H}_{v}^{t}=\mathbf{X}_{v}-\mathbf{X}_{v} \mathbf{S}_{v}^{t+1}+\frac{\mathbf{W}_{v}^{t}}{\tau^{t}}, 
%\end{equation}
%$\mathbf{H}^{t}$ is constructed by vertically concatenating the matrices $\left\{\mathbf{H}_{v}\right\}, v = 1,\cdots,V$. As it is suggested in~\cite{liu2012robust}, (\ref{eq:upeem}) has a closed-form solution:
\begin{equation}
\mathbf{E}_{i}^{t+1}=\left\{\begin{array}{ll}
\frac{\left\|\mathbf{H}_{i}^{t}\right\|_{2}-\frac{1}{\tau^{t}}}{\left\|\mathbf{H}_{i}^{t}\right\|_{2}} \mathbf{H}_{i}^{t}, & \text { if } \frac{1}{\tau^{t}}<\left\|\mathbf{H}_{i}^{t}\right\|_{2} \\
0, & \text { otherwise }
\end{array}\right.
\label{eq:upe}
\end{equation}
where $\mathbf{H}_{i}^{t}$ and  $\mathbf{E}_{i}^{t+1}$ denote the $i$-th column of $\mathbf{H}^{t}$ and $\mathbf{E}^{t+1}$, respectively.

\textbf{Solving $\mathbf{M}$-subproblem:} 
The optimization subproblem with respect to $\mathbf{M}$ is:
\begin{equation}\label{eq:a}
\begin{aligned}
&\min \limits_{\mathbf{M}}\sum \limits_{v=1}^{V} \mu \operatorname{tr}(\mathbf{S}_{v}^{{t+1}^{\mathrm{T}}} \mathbf{L}\mathbf{S}_{v}^{t+1})+\lambda\|\mathbf{M}\|_{\mathrm{F}}^{2} \\
&~\text {s. t.}~~\mathbf{M}^{\mathrm{T}} \mathbf{1}=\mathbf{1}, \mathbf{0} \leq \mathbf{M} \leq \mathbf{1}.
\end{aligned}
\end{equation}
%where $\mathbf{L}=\mathbf{D}-(\mathbf{M}+\mathbf{M}^{\mathrm{T}})/2$. 

According to ~(\ref{eq:mgm}) and ~(\ref{eq:trmgm}), ~(\ref{eq:a}) can be divided into $i$ subproblems:
\begin{equation}\label{eq:a1}
\begin{aligned}
&\min \limits_{\mathbf{M}_{i}}\sum \limits_{j=1}^{N} \sum \limits_{v=1}^{V} \frac{\mu}{2}\|(\mathbf{S}_{i}^{t+1})_{v}-(\mathbf{S}_{j}^{t+1})_{v}\|_{2}^{2} m_{i,j}+\lambda \sum \limits_{j=1}^{N} m_{i ,j}^{2} \\
&~\text {s. t.}~\mathbf{M}_{i}^{\mathrm{T}} \mathbf{1}=1, \mathbf{0} \leq \mathbf{M}_{i} \leq \mathbf{1}.
\end{aligned}
\end{equation}
where $\mathbf{M}_{i}$ is the  $i$-th column of matrix $\mathbf{M}$ and $(\mathbf{S}_{i})_{v}$ is the  $i$-th column of matrix $\mathbf{S}_{v}$. Following~\cite{kang2019robust,chen2021low} that apply the adaptive neighbor scheme to keep $K$ largest values constant in $\mathbf{M}_{i}$ and the rest become zero, each subproblem has the solution:
\begin{equation}
\left\{\begin{array}{l}
m_{i j}^{l+1}=\frac{p_{i, K+1}-p_{i j}}{K p_{i, K+1}-\sum_{k=1}^{K} p_{i,k}}, j \leq K\\
m_{i j}^{l+1}=0,j>K.
\end{array}\right.
\label{eq:mk}
\end{equation}
where $p_{i,j}=\sum_{v=1}^{V}\|(\mathbf{s}_{i}^{l+1})_{v}-(\mathbf{s}_{j}^{l+1})_{v}\|_{2}^{2}$ and $\mathbf{p}_{i} \in \mathbb{R}^{N \times 1}$.

\textbf{Updating $\mathcal{W}$, $\mathcal{Y}$ and $\tau$:} 
By keeping other variables, i.e., $\mathcal{Z}$, $\mathcal{S}$ and $\mathbf{E}$ are fixed, the Lagrangian multipliers $\mathcal{W}, \mathcal{Y}$ and the penalty parameter $\tau$ are updated as follows:
\begin{equation}
\begin{aligned}
\mathbf{W}_{v}^{t+1} &=\mathbf{W}_{v}^{t}+\tau^{t}\left(\mathbf{X}_{v}-\mathbf{X}_{v} \mathbf{S}_{v}^{t+1}-\mathbf{E}_{v}^{t+1}\right), \\
\mathcal{Y}^{t+1} &=\mathcal{Y}^{t}+\tau^{t}\left(\mathcal{Z}^{t+1}-\mathcal{S}^{t+1}\right), \\
\tau^{t+1} &=\min \left\{\beta  \tau^{t}, \tau_{max}\right\},
\end{aligned}
\label{eq:upwy}
\end{equation}
where $\beta>1$ is utilized to accelerate the convergence, and $\tau_\text{max}$ is the predefined maximum value of $\tau$. %The overall procedure to solve the multi-view clustering optimization problem is summarized in Algorithm 2 in our supplement \ref{algo:2}.  
The affinity matrix $\mathbf{M}$, which will be put into spectral clustering algorithm to yield the final clustering results, can be calculated by $\mathbf{M}=1/V \sum_{v=1}^{V}(|\mathbf{Z}_{v}|+|\mathbf{Z}_{v}^{\mathrm{T}}|).$
% \begin{equation}
% \mathbf{M}=\frac{1}{V} \sum_{v=1}^{V}(|\mathbf{Z}_{v}|+|\mathbf{Z}_{v}^{\mathrm{T}}|).
% \label{eq:affinity}
% \end{equation}
%where $V$ is the number of views, and $\mathbf{Z}_{v}~(v=1,\cdots,V)$ represents the $v$-th view's learned representation matrix. %The affinity matrix $\mathbf{M}$ can be put into the clustering.

\begin{center}
		\begin{algorithm}[tb]
			\caption{TOMD for low rank multi-view clustering}
			\begin{algorithmic}
				%\STATE \textbf{Input}: Multi-view data $\{\mathbf{X}_{v}\}, v=1,\cdots,V$, parameter $\mu$ and the number of nearest neighbors $K$, $\text{iter}_\text{max}=150$, the predefined size of reshaped 4th-order tensor $(N_1,N_2,N_3,V)$.
				\STATE \textbf{Input}: Multi-view data $\{\mathbf{X}_{v}\}, v=1,\cdots,V$, $\mu$ and $K$, $\text{iter}_\text{max}=150$, and $N_1,N_2,N_3,V$.
				\STATE \textbf{Initialization}: $\mathcal{S}, \mathcal{Z}, \mathbf{E}, \mathbf{M},\mathcal{W}, \mathcal{Y}$ initialized to $\mathbf{0}$, $\tau=1$, $\beta=1.5$, $\text{tol}=10^{-7}$, iteration $t=1$.
				\WHILE{$t \leq \text{iter}_\text{max}$}
				\STATE  $\tilde{\mathcal{S}}^{t}=\Phi((\mathcal{S}^{t}-\frac{\Pi^{t}}{\tau^{t}}),[N_1,N_2,N_3,V])$;
				\STATE  Obtain $\{\tilde{\mathcal{G}}^{(n)}\}_{n=1}^{5},\{\tilde{\mathbf{U}}^{(n)}\}_{n=1}^{4}$ by TOMD-ALS and update $\mathcal{Z}^{t+1}$ by (\ref{eq:upzpro}); %by  putting $\tilde{\mathcal{S}}^{t}$ into Algorithm \ref{algo:1};
				%\STATE Update $\mathcal{Z}^{t+1}$ by (\ref{eq:upzpro});
				\FOR{$v=1:V$}
				\STATE Update $\mathbf{S}_{v}^{t+1}$;
				\ENDFOR
				\STATE Update $\mathbf{E}^{t+1}$ and $\mathbf{M}^{t+1}$ by (\ref{eq:upe}) and (\ref{eq:mk}), respectively; 
				\STATE Update $\mathcal{W}^{t+1}$, $\mathcal{Y}^{t+1}$ and $\tau^{t+1}$ by ~(\ref{eq:upwy});	
				\STATE Check convergence conditions:  
				\STATE $\max(\|\mathbf{X}_{v}-\mathbf{X}_{v} \mathbf{S}_{v}^{t+1}-\mathbf{E}_{v}^{t+1}\|_{\infty} ,\|\mathcal{Z}^{t+1}-\mathcal{S}^{t+1}\|_{\infty}) \leq \text{tol}$;
				\STATE t=t+1;
				%  \STATE \textbf{if} $\max(\|\mathbf{X}_{v}-\mathbf{X}_{v} \mathbf{S}_{v}^{t+1}-\mathbf{E}_{v}^{t+1}\|_{\infty} ,\|\mathcal{Z}^{t+1}-\mathcal{S}^{t+1}\|_{\infty}) \leq \text{tol}$
				% \STATE \quad break;
				% \STATE \textbf{endif} 						
				\ENDWHILE
				\STATE $\textbf{Output:}$ Affinity matrix $\mathbf{M}^{t+1}$.
			\end{algorithmic}
			\label{algo:2}
		\end{algorithm}
\end{center}

\subsection{Computational Complexity}
To analyze the computation complexity of TOMD-MVC, we assume that $R=R_{1}=\cdots=R_{4}=D_{1}=\cdots=D_{6}$ for convenience of analysis. At each iteration, it takes $\mathcal{O}\left(IN_{1}N_{2}N_{3}VR^{4}\right)$ to update the reshaped tensor $\mathcal{Z}\in\mathbb{R}^{N_{1}\times N_{2}\times N_{3}\times V}$, where $I$ denotes number of iterations of TOMD-ALS. For $\mathcal{S}$-subproblem, we need $\mathcal{O}\left(V N^{3}\right)$ to compute its closed-form. Updating $\mathbf{E}$ and $\mathbf{M}$ cost $\mathcal{O}\left(V N^{2}\right)$ and $\mathcal{O}\left(N^{2}\right)$, respectively. Therefore, the computation in each iteration is $\mathcal{O}\left(V(IN_{1}N_{2}N_{3}R^{4}+N^{3}+N^{2})+N^{2}\right)$. 

In addition, after we get the affinity matrix, we adopt the spectral clustering method to obtain the final memberships of data. Considering the spectral clustering's computational complexity is $\mathcal{O}\left(N^{3}\right)$~\cite{wu2019essential}, the overall complexity is $\mathcal{O}\left(T V(IN_{1}N_{2}N_{3}R^{4}+N^{3}+N^{2})+L N^{2}+ N^{3}\right)$, where $T$ is the iteration number of TOMD-MVC.

\section{Experiments}
\label{sec:6}
In this section, image reconstruction experiments have been employed to indicate our proposed tensor network's advantages in depicting low-rank properties. The clustering performance of our proposed method has also been demonstrated over six multi-view datasets. 

\subsection{Low-Rank Analysis}
%To illustrate the superiority of our proposed tensor network, we conduct the image reconstruction experiments on several real word images\footnote{http://sipi.usc.edu/database/database.php?volume=misc}, including RGB images: "House", "Peppers", "Airplane (F-16)", "Sailboat on lake" with size $256 \times 256 \times 3$ and gray image "Airplane" with size $256 \times 256$. Since the TOMD has been utilized for 4th-order tensors, the gray forms of these images are transformed into $16 \times 16 \times 16 \times 16$ tensors.
To illustrate the superiority of our proposed tensor network, we conduct the image reconstruction experiments on several real word images\footnote{http://sipi.usc.edu/database/database.php?volume=misc}: "House", "Peppers", "Airplane (F-16)", "Sailboat on lake", and "Airplane". Since the TOMD has been utilized for 4th-order tensors, the gray forms of these images are further transformed into $16 \times 16 \times 16 \times 16$ tensors.

%In this section, the effectiveness and low-rank property of the algorithms for image reconstruction have been compared with each other. The tensors are performed by Tucker-ALS, Tucker and tensor ring decomposition ALS (TuTR-ALS), O-Minus-ALS (Ominus-ALS) and TOMD-ALS algorithms with the same Relative standard error (RSE, whose definition is
% ($\operatorname{RSE}=\|\mathcal{X}_{\text{new}}-\mathcal{X}\|_{\mathrm{F}}/\|\mathcal{X}\|_{\mathrm{F}}$)
% (RSE, whose definition is $\operatorname{RSE}=\|\mathcal{X}_{\text{new}}-\mathcal{X}\|_{\mathrm{F}}/\|\mathcal{X}\|_{\mathrm{F}}$, where $\mathcal{X}_{\text{new}}$ is the recovered tensor and $\mathcal{X}$ is the original one.)
%In this section, the tensors are performed by Tucker-ALS, Tucker and tensor ring decomposition ALS (TuTR-ALS), O-Minus-ALS (Ominus-ALS) and TOMD-ALS algorithms with the same Relative standard error (RSE, whose definition is presented in (\ref{eq:rse})). 
In this section, the tensors are performed by Tucker-ALS, Tucker and tensor ring decomposition ALS (TuTR-ALS), O-Minus-ALS (Ominus-ALS) and TOMD-ALS algorithms with the same Relative standard error (RSE) \cite{yuan2018higher}. 
Beyond that, the storage costs of the different decomposition have been calculated and presented in Table~\ref{tab:comimage}. In addition to the time needed for pre-processing the images, reconstructing the images, and calculating the storage costs, the running time in our experiments only contains the main body in decomposition, which is also illustrated in Table~\ref{tab:comimage}. Considering that the speed of the iterative algorithm primarily affects the running time of those methods, we set the number of iterations for all methods to 500 in our experiment.
% \begin{equation}
% \operatorname{RSE}=\frac{\|\mathcal{X}_{\text{new}}-\mathcal{X}\|_{\mathrm{F}}}{\|\mathcal{X}\|_{\mathrm{F}}},
% \label{eq:rse}
% \end{equation}
% where $\mathcal{X}_{\text{new}}$ is the recovered tensor and $\mathcal{X}$ is the original one.

As results shown in Table~\ref{tab:comimage}, under the same conditions, we can conclude that the running time of TOMD is in the middle, but it has the lowest storage cost and can better exploit the low rank information compared to the other decompositions. 

\begin{table}[htbp]
\begin{center}
\caption{Comparison results about Tucker-ALS, TuTR-ALS, Ominus-ALS and TOMD-ALS algorithms based image reconstruction experiments, and the values under the datasets' name are the predefined RSE.}
\label{tab:comimage}
\scalebox{1.1}{
\begin{tabular}{{ccccc}}
\hline		
Dataset &Method &Time(s) &Storage Cost 
\\
\hline			
	   	&Tucker-ALS 	& 17.10	&2473	\\
House	&TuTR-ALS 	& 20.21	&1872	\\
(0.12)   	&Ominus-ALS	& \textbf{11.34}	&1545	\\
        		&TOMD-ALS	& 17.22	&\textbf{1220}	\\
\hline			
		&Tucker-ALS 	& 80.74	&11904	\\
Peppers	&TuTR-ALS 	& \textbf{40.51}	&4898	\\
(0.12)	&Ominus-ALS	& 45.93	&4825	\\
        		&TOMD-ALS	& 47.48	&\textbf{3072}	\\
\hline			
		&Tucker-ALS 	& 55.66	&4608	\\
F-16		&TuTR-ALS	& 29.12	&2340	\\
(0.12)	&Ominus-ALS	& \textbf{12.41}	&1801	\\
        		&TOMD-ALS	& 17.22	&\textbf{1484}	\\
\hline			
		&Tucker-ALS  	& 85.83	&10640	\\
Sailboat 	&TuTR-ALS 	& \textbf{45.03}	&5820	\\
(0.12)	&Ominus-ALS	& 51.66	&6361	\\
        		&TOMD-ALS	& 47.62	&\textbf{4894}	\\
\hline			
		&Tucker-ALS  	& \textbf{11.30}	&2849	\\
Airplane	&TuTR-ALS 	& 30.22	&2080	\\
(0.06)	&Ominus-ALS	& 12.49	&2507	\\
        		&TOMD-ALS	& 26.81	&\textbf{1516}	\\
\hline
\end{tabular}
}
\end{center}
\end{table}

\subsection{Experimental Settings For Multi-view Clustering}

\subsubsection{Datasets}
The performance of TOMD-MVC has been investigated over six multi-view datasets, Yale\footnote{http://cvc.cs.yale.edu/cvc/projects/yalefaces/yalefaces.html}, MSRCV1\footnote{http://research.microsoft.com/en-us/projects/objectclassrecognition/}, ExtendYaleB~\cite{wang2017exclusivity}, ORL\footnote{http://www.uk.research.att.com/facedatabase.html}, Reuters\footnote{http://lig-membres.imag.fr/grimal/data.html} and Handwritten~\cite{chen2021smoothed}, and their statistic information has been summarized in Table \ref{tab:data}.
%The performance of TOMD-MVC has been investigated over six multi-view datasets, Yale\footnote{http://cvc.cs.yale.edu/cvc/projects/yalefaces/yalefaces.html}, MSRCV1\footnote{http://research.microsoft.com/en-us/projects/objectclassrecognition/}, ExtendYaleB~\cite{wang2017exclusivity}, ORL\footnote{http://www.uk.research.att.com/facedatabase.html}, Reuters\footnote{http://lig-membres.imag.fr/grimal/data.html} and Handwritten~\cite{chen2021smoothed}, which cover four different applications, including text clustering, generic object clustering, facial clustering and digit clustering. We have summarized their statistic information in Table \ref{tab:data} and briefly introduce these data sets as follows.
\begin{table*}[htbp]
\centering
\caption{Statistics of different datasets.}
	\scalebox{0.7}{
	\begin{tabular}{{ccccccc}}
		\hline
		Datasets 		&Feature($C_{1},\cdots,C_{v}$) 	&Samples(N) 	&Views(V)  	&Clusters 	&Objective 	&($N_{1},N_{2},N_{3},V$)\\
		\hline
		Yale			&(4096,3304,6750)  			&165  		&3   			&15    	&Face 		&(165,15,11,3)\\
		MSRCV1		&(1302,48,512,100,256,210)  	&210  		&6 			&7  		&Object 		&(210,15,14,6)\\
		ExtendYaleB 	&(2500,3304,6750)  			&650  		&3 			&10  		&Face 		&(50,13,650,3)\\
		ORL     		&(4096,3304,6750)    		&400  		&3  			&40 		&Face 		&(400,20,20,3)\\
		Reuters     	&(4819,4810,4892,4858,4777)  &1200    &5      &6  &Text       &(1200,20,60,5)\\
		Handwritten      &(240,76,216,47,64,6) 		&2000 	&6 	&10 	&Digit	&(200,10,200,60)\\
		\hline	
	\end{tabular}
	}
\label{tab:data}
\end{table*}

% \textbf{Yale} face dataset contains 165 grayscale images of 15 individuals. There are 11 images per subject, one per different facial expression or configuration. \textbf{MSRCV1} consists of 210 image samples collected from 7 clusters with 6 views, including CENT, CMT, GIST, HOG, LBP, and SIFT. \textbf{ExtendYaleB} contains 650 face images of 10 individuals, each of which has approximately 65 frontal images under different lighting conditions. \textbf{ORL} contains 400 face images of 40 distinct subjects. Each subject has 10 different face images taken at different times, changing with the lighting, facial expressions and facial details. \textbf{Reuters} contains 1200 documents, each of which described with five languages, including English, French, German, Italian, and Spanish. \textbf{Handwritten} contains 2000 images of digits 0 to 9 with pixel averages, Fourier coefficient, profile correlations, Zernike moments, Karhunen coefficient, and morphological six features.

%In addition, we compare our TOMD-MVC model with two graph-based methods, GMC~\cite{wang2019gmc}, GFSC~\cite{kang2020multi}, and ten multi-view clustering methods,  LMSC~\cite{zhang2017latent}, SFMC~\cite{li2020multiview}, EOMSC-CA~\cite{liu2022efficient}, LTMSC~\cite{zhang2015low}, t-SVD-MSC~\cite{xie2018unifying}, ETLMSC~\cite{wu2019essential}, UGLTL~\cite{wu2020unified}, HLR-M$^{2}$VC~\cite{xie2020hyper}, COMSC~\cite{liu2021multiview}, TBGL~\cite{xia2022tensorized}, to verify its superiority. 
\subsubsection{Competitors} We compare our TOMD-MVC model with six matrix based methods, GMC~\cite{wang2019gmc}, GFSC~\cite{kang2020multi}, LMSC~\cite{zhang2017latent}, SFMC~\cite{li2020multiview}, EOMSC-CA~\cite{liu2022efficient}, COMSC~\cite{liu2021multiview}, and six multi-view clustering methods, LTMSC~\cite{zhang2015low}, t-SVD-MSC~\cite{xie2018unifying}, ETLMSC~\cite{wu2019essential}, UGLTL~\cite{wu2020unified}, HLR-M$^{2}$VC~\cite{xie2020hyper}, TBGL~\cite{xia2022tensorized}, to verify its superiority. 
\subsubsection{Evaluation Metrics}
In our experiments, six commonly used metrics are applied to quantitatively evaluate the clustering performance, namely F-score, precision (P), recall (R), normalized mutual information (NMI), adjusted rand index (AR), and accuracy (ACC). In addition, the higher values of these metrics indicate better clustering performance. Further detailed information please refer to~\cite{lin2018multi,zhang2021joint,huang2019ultra,schutze2008introduction,zhong2003unified}. 

\subsection{Clustering Performance Comparison}
\subsubsection{Compared with state-of-the-art methods}
%All detailed clustering results on five datasets are presented in Table~\ref{tab:yale}-\ref{tab:hand}. Since the different initializations may lead to different results, we run 10 trials for each experiment and present their average performance with standard deviations, i.e., mean(standard deviation). The best results are highlighted in \textbf{boldface} and the second-best results are $\underline{\text{underlined}}$. From those experiment results, we can obtain the following conclusions.
All detailed clustering results on six data sets are presented in Table~\ref{tab:yale}-\ref{tab:hand}. Since the different initializations may lead to different results, we run 10 trials for each experiment and present their average performance with standard deviations, i.e., mean(standard deviation). The best results are highlighted in \textbf{boldface} and the second-best results are $\underline{\text{underlined}}$. %From those experiment results, we can obtain the following conclusions.

LTMSC, t-SVD-MSC, ETLMSC, UGLTL, HLR-M$^{2}$VC, and TBGL are all tensor-based methods, whose clustering performance is almost higher than matrix based methods. Therefore, employing the tensor based low rank constraints on self-representation tensor is a better way to capture the higher-order correlations. Nevertheless, our proposed TOMD-MVC can still achieve the best results among these tensor-based methods under all six evaluation metrics. Specifically, the results of our proposed model on Yale, ORL and Reuters data sets, are $1.5\%$,  $1.13\%$, $6.57\%$ higher than the second-best method, respectively. Especially on the ExtendYaleB dataset, the improvement of TOMD-MVC is at least $30\%$ higher than the second-best performance of $\text{t-SVD-MSC}$. Based on this, we can demonstrate the superiority of our proposed tensor network in low rank multi-view clustering.

\begin{table*}[htbp]
\begin{center}
\caption{Clustering results on Yale dataset. For TOMD-MVC, we set $K=10$ and $\mu=1$.}
\label{tab:yale}
\scalebox{0.6}{
\begin{tabular}{{cccccccccc}}
\hline		
&Method &F-score &Precision &Recall &NMI &AR &ACC
\\
	\hline
		&GMC 		&0.446(0.000)		&0.378(0.000)	&0.544(0.000)	&0.668(0.000)	&0.403(0.000)	&0.618(0.000)\\
		&GFSC 		&0.433(0.028)		&0.384(0.040)	&0.497(0.016)	&0.647(0.015)	&0.391(0.032)	&0.621(0.029)\\
		&LMSC 		&0.519(0.003)		&0.475(0.006)	&0.572(0.000)	&0.717(0.002)	&0.484(0.004)	&0.679(0.007)\\
		&SFMC		&0.480(0.000)		&0.438(0.000)	&0.532(0.000)	&0.663(0.000)	&0.443(0.000)	&0.618(0.000)\\		
		&EOMSC-CA &0.469(0.000)		&0.418(0.000)	&0.533(0.000)	&0.654(0.000)	&0.430(0.000)	&0.648(0.000)\\
 		&LTMSC 		&0.620(0.009)		&0.599(0.011)	&0.643(0.006)	&0.764(0.006)	&0.594(0.009)	&0.736(0.004)\\
		&t-SVD-MSC 	&\underline{0.902(0.066)}		&\underline{0.891(0.075)}	&\underline{0.915(0.058)}	&\underline{0.946(0.036)}	&\underline{0.896(0.071)}	&\underline{0.934(0.053)}\\
		&ETLMSC	&0.542(0.055)		&0.509(0.053)	&0.580(0.061)	&0.706(0.043)	&0.510(0.059)	&0.654(0.044)\\
		&UGLTL 		&0.867(0.047)		&0.852(0.057)	&0.883(0.039)	&0.931(0.023)	&0.858(0.050)	&0.912(0.041)\\
		&HLR-M$^{2}$VC &0.695(0.010)	&0.673(0.011)	&0.718(0.010)	&0.817(0.006)	&0.674(0.011)	&0.772(0.012)\\
		&COMSC		&0.627(0.000)		&0.607(0.000)	&0.953(0.000)	&0.762(0.000)	&0.602(0.000)	&0.739(0.000)\\
		&TBGL		&0.587(0.000)		&0.556(0.000)	&0.623(0.000)	&0.739(0.000)	&0.559(0.000)	&0.703(0.000)\\
		&TOMD-MVC	&\textbf{0.916(0.013)} &\textbf{0.914(0.014)} &\textbf{0.918(0.013)} &\textbf{0.949(0.008)} &\textbf{0.910(0.014)} &\textbf{0.959(0.006)}\\
\hline
\end{tabular}
}
\end{center}
%\label{tab:yale}
\end{table*}

\begin{table*}[htbp]
\begin{center}
\caption{Clustering results on MSRCV1 dataset. For TOMD-MVC, we set $K=5$ and $\mu=50$.}
\label{tab:msrcv1}
\scalebox{0.6}{
\begin{tabular}{{cccccccccc}}
\hline		
&Method &F-score &Precision &Recall &NMI &AR &ACC
\\
\hline
		&GMC 		&0.824(0.000)	&0.812(0.000)	&0.837(0.000)	&0.846(0.000)	&0.795(0.000)	&0.910(0.000)\\
		&GFSC 		&0.609(0.050)	&0.581(0.065)	&0.642(0.031)	&0.661(0.042)	&0.542(0.062)	&0.730(0.061)\\
		&LMSC 		&0.656(0.081)	&0.646(0.081)	&0.667(0.080)	&0.677(0.065)	&0.600(0.094)	&0.783(0.086)\\
		&SFMC   &0.763(0.000)	&0.701(0.000)	&0.836(0.000)	&0.790(0.000)	&0.720(0.000)	&0.838(0.000)\\
		&EOMSC-CA &0.813(0.000)	&0.787(0.000)	&0.840(0.000)	&0.837(0.000)	&0.781(0.000)	&0.876(0.000)\\
		&LTMSC  		&0.727(0.001)	&0.714(0.001)	&0.742(0.000)	&0.750(0.000)	&0.682(0.001)	&0.829(0.000)\\
		&t-SVD-MSC 	&\underline{0.962(0.000)}	&\underline{0.961(0.000)}	&\underline{0.963(0.000)}	&\underline{0.960(0.000)}	&\underline{0.955(0.000)}	&\underline{0.981(0.000)}\\
		&ETLMSC	&0.934(0.079)	&0.924(0.099)	&0.946(0.056)	&0.946(0.055)	&0.923(0.092)	&0.950(0.077)\\	
		&UGLTL 	& \textbf{1.000(0.000)} &\textbf{1.000(0.000)}	 &\textbf{1.000(0.000)} &\textbf{1.000(0.000)} &\textbf{1.000(0.000)} &\textbf{1.000(0.000)}\\
		&HLR-M$^{2}$VC &0.990(0.000)	&0.990(0.000)	&0.990(0.000)	&0.989(0.000)	&0.989(0.000)	&0.995(0.000)\\
		&COMSC &0.861(0.000)	&0.856(0.000)	&0.961(0.000)	&0.863(0.000)	&0.838(0.000)	&0.929(0.000)\\
		&TBGL	& \textbf{1.000(0.000)} &\textbf{1.000(0.000)}	 &\textbf{1.000(0.000)} &\textbf{1.000(0.000)} &\textbf{1.000(0.000)} &\textbf{1.000(0.000)}\\
		&TOMD-MVC 	& \textbf{1.000(0.000)} &\textbf{1.000(0.000)}	 &\textbf{1.000(0.000)} &\textbf{1.000(0.000)} &\textbf{1.000(0.000)} &\textbf{1.000(0.000)}\\
\hline
\end{tabular}
}
\end{center}
\end{table*}

\begin{table*}[htbp]
\begin{center}
\caption{Clustering results on ExtendYaleB dataset. For TOMD-MVC, we set $K=15$ and $\mu=50$.}
\label{tab:yaleb}
\scalebox{0.6}{
\begin{tabular}{{cccccccccc}}
\hline		
&Method &F-score &Precision &Recall &NMI &AR &ACC
\\
\hline
		&GMC&0.256(0.000)	&0.201(0.000)	&0.351(0.000)	&0.405(0.000)	&0.149(0.000)	&0.386(0.000)\\
		&GFSC &0.204(0.006)	&0.161(0.004)	&0.282(0.024)	&0.327(0.023)	&0.090(0.005)	&0.348(0.025)\\
		&LMSC &0.352(0.001)	&0.309(0.001)	&0.410(0.001)	&0.513(0.001)	&0.270(0.001)	&0.530(0.001)\\
		&SFMC &0.290(0.000)	&0.197(0.000)	&0.551(0.000)	&0.494(0.000)	&0.170(0.000)	&0.509(0.000)\\
		&EOMSC-CA &0.202(0.000)	&0.169(0.000)	&0.252(0.000)	&0.250(0.000)	&0.095(0.000)	&0.289(0.000)\\
		&LTMSC 	&0.323(0.005)	&0.296(0.006)	&0.355(0.004)	&0.456(0.004)	&0.242(0.006)	&0.455(0.002)\\
		&t-SVD-MSC &0.483(0.005)	&0.456(0.007)	&0.513(0.004)	&0.618(0.003)	&0.422(0.006)	&0.577(0.001)\\
		&ETLMSC&0.262(0.017)	&0.257(0.017)	&0.590(0.008)	&0.307(0.021)	&0.179(0.019)	&0.325(0.011)\\
		&UGLTL 	&0.406(0.038)	&0.395(0.036)	&0.417(0.041)	&0.503(0.044)	&0.339(0.042)	&0.463(0.037)\\
		&HLR-M$^{2}$VC &\underline{0.575(0.000)}	&\underline{0.553(0.000)}	&0.599(0.000)	&\underline{0.706(0.000)}	&\underline{0.527(0.000)}	&\underline{0.673(0.000)}\\
		&COMSC &0.453(0.000)	&0.414(0.000)	&\underline{0.881(0.000)}	&0.596(0.000)	&0.387(0.000)	&0.616(0.000)\\
		&TBGL	&0.230(0.000)	&0.156(0.000)	&0.437(0.000)	&0.382(0.000)	&0.100(0.000)	&0.403(0.000)\\
		&TOMD-MVC  	&\textbf{0.981(0.001)} &\textbf{0.981(0.001)}	&\textbf{0.981(0.001)}	&\textbf{0.982(0.001)}	&\textbf{0.979(0.001)}	&\textbf{0.990(0.001)}\\

\hline
\end{tabular}
}
\end{center}
\end{table*}

\begin{table*}[htbp]
\begin{center}
\caption{Clustering results on ORL dataset. For TOMD-MVC, we set $K=10$ and $\mu=30$.}
\label{tab:orl}
\scalebox{0.6}{
\begin{tabular}{{cccccccccc}}
\hline		
&Method &F-score &Precision &Recall &NMI &AR &ACC
\\
\hline
	&GMC  		&0.368(0.000)	&0.239(0.000)	&0.805(0.000)	&0.861(0.000)	&0.345(0.000)	&0.660(0.000)\\
	&GFSC 		&0.501(0.053)	&0.413(0.063)	&0.644(0.035)	&0.828(0.019)	&0.487(0.055)	&0.636(0.046)\\
	&LMSC 		&0.776(0.016)	&0.717(0.024)	&0.847(0.013)	&0.932(0.005)	&0.771(0.017)	&0.823(0.016)\\	
	&SFMC &0.702(0.000)	&0.586(0.000)	&0.876(0.000)	&0.908(0.000)	&0.694(0.000)	&0.775(0.000)\\
	&EOMSC-CA &0.461(0.000)	&0.365(0.000)	&0.627(0.000)	&0.791(0.000)	&0.446(0.000)	&0.613(0.000)\\
	&LTMSC 		&0.766(0.025)	&0.725(0.028)	&0.813(0.023)	&0.920(0.010)	&0.760(0.026)	&0.817(0.021)\\
	&t-SVD-MSC 	&0.987(0.017)	&0.979(0.027)	&0.996(0.006)	&0.997(0.003)	&0.987(0.017)	&0.987(0.017)\\
	&ETLMSC	&0.928(0.027)	&0.894(0.036)	&0.966(0.019)	&0.982(0.009)	&0.927(0.028)	&0.929(0.025)\\
	&UGLTL 	&0.948(0.017)	&0.932(0.022)	&0.965(0.013)	&0.985(0.005)	&0.947(0.017)	&0.954(0.016)\\
	&HLR-M$^{2}$VC &\underline{0.991(0.015)}	&\underline{0.984(0.026)}	&\underline{0.997(0.005)}	&\underline{0.998(0.003)}	&\underline{0.990(0.016)}	&\underline{0.991(0.015)}\\
	&COMSC &0.770(0.000)	&0.730(0.000)	&0.989(0.000)	&0.913(0.000)	&0.764(0.000)	&0.835(0.000)\\
	&TBGL	&0.611(0.000)	&0.470(0.000)	&0.871(0.000)	&0.895(0.000)	&0.599(0.000)	&0.848(0.000)\\
	&TOMD-MVC 	&\textbf{1.000(0.000)} &\textbf{1.000(0.000)} 	&\textbf{1.000(0.000)} &\textbf{1.000(0.000)} 	&\textbf{1.000(0.000)} 	&\textbf{1.000(0.000)} \\
\hline
\end{tabular}
}
\end{center}
\end{table*}

\begin{table*}[htbp]
\begin{center}
\caption{Clustering results on Reuters dataset. For TOMD-MVC, we set $K=20$ and $\mu=50$.}
\label{tab:reuters}
\scalebox{0.6}{
\begin{tabular}{{cccccccccc}}
\hline		
&Method &F-score &Precision &Recall &NMI &AR &ACC
\\
\hline
	&GMC  		&0.283(0.000)	&0.167(0.000)	&0.927(0.000)	&0.075(0.000)	&0.003(0.000)	&0.187(0.000)\\
	&GFSC 		&0.328(0.010)	&0.254(0.020)	&0.475(0.055)	&0.226(0.024)	&0.143(0.026)	&0.375(0.045)\\
	&LMSC 		&0.407(0.000)	&0.374(0.000)	&0.447(0.000)	&0.342(0.000)	&0.276(0.000)	&0.562(0.000)\\
	&SFMC &0.284(0.000)	&0.166(0.000)	&0.966(0.000)	&0.019(0.000)	&0.001(0.000)	&0.182(0.000)\\
	&EOMSC-CA &0.312(0.000)	&0.291(0.000)	&0.335(0.000)	&0.223(0.000)	&0.163(0.000)	&0.427(0.000)\\
	&LTMSC 		&0.304(0.000)	&0.253(0.000)	&0.380(0.000)	&0.196(0.001)	&0.130(0.000)	&0.382(0.001)\\
	&t-SVD-MSC 	&\underline{0.904(0.000)}	&\underline{0.902(0.000)}	&\underline{0.907(0.000)}	&\underline{0.885(0.000)}	&\underline{0.885(0.000)}	&\underline{0.950(0.000)}\\
	&ETLMSC	&0.898(0.114)	&0.891(0.128)	&0.907(0.098)	&0.896(0.087)	&0.877(0.138)	&0.920(0.118)\\
	&UGLTL 	&0.934(0.000)	&0.933(0.000)	&0.934(0.000)	&0.914(0.000)	&0.920(0.000)	&0.966(0.000)\\
	&HLR-M$^{2}$VC &0.714(0.001)	&0.704(0.001)	&0.724(0.001)	&0.708(0.001)	&0.656(0.001)	&0.831(0.001)\\
	&COMSC &0.407(0.000)	&0.349(0.000)	&0.764(0.000)	&0.330(0.000)	&0.264(0.000)	&0.547(0.000)\\
	&TBGL	&0.284(0.000)	&0.167(0.000)	&0.951(0.000)	&0.029(0.000)	&0.002(0.000)	&0.179(0.000)\\
	&TOMD-MVC 	&\textbf{0.964(0.013)}	&\textbf{0.963(0.014)}	&\textbf{0.966(0.012)}	&\textbf{0.957(0.016)}	&\textbf{0.957(0.016)}	&\textbf{0.981(0.007)} \\
\hline
\end{tabular}
}
\end{center}
\end{table*}

\begin{table*}[htbp]
\begin{center}
\caption{Clustering results on Handwritten dataset. For TOMD-MVC, we set $K=20$ and $\mu=40$.}
\label{tab:hand}
\scalebox{0.6}{
\begin{tabular}{{cccccccccc}}
\hline		
&Method &F-score &Precision &Recall &NMI &AR &ACC\\
\hline
	&GMC  		&0.854(0.000)	&0.793(0.000)	&0.925(0.000)	&0.910(0.000)	&0.836(0.000)	&0.857(0.000)\\
	&GFSC 		&0.630(0.049)	&0.573(0.052)	&0.700(0.054)	&0.715(0.026)	&0.584(0.056)	&0.714(0.053)\\
	&LMSC 		&0.818(0.001)	&0.815(0.001)	&0.820(0.001)	&0.816(0.001)	&0.797(0.001)	&0.901(0.000)\\
	&SFMC &0.868(0.000)	&0.820(0.000)	&0.920(0.000)	&0.892(0.000)	&0.852(0.000)	&0.881(0.000)\\
	&EOMSC-CA &0.710(0.000)	&0.613(0.000)	&0.842(0.000)	&0.775(0.000)	&0.672(0.000)	&0.725(0.000)\\
	&LTMSC 		&0.818(0.016)	&0.815(0.016)	&0.821(0.015)	&0.833(0.010)	&0.797(0.017)	&0.896(0.012)\\
	&t-SVD-MSC 	&\underline{0.999(0.000)}	&\underline{0.999(0.000)}	&\underline{0.999(0.000)}	&\underline{0.999(0.000)}	&\underline{0.999(0.000)}	&\underline{1.000(0.000)}\\
	&ETLMSC	&0.919(0.071)	&0.880(0.107)	&0.965(0.027)	&0.964(0.030)	&0.909(0.080)	&0.902(0.089)\\
	&UGLTL 	&\textbf{1.000(0.000)} &\textbf{1.000(0.000)} 	&\textbf{1.000(0.000)} &\textbf{1.000(0.000)} 	&\textbf{1.000(0.000)} 	&\textbf{1.000(0.000)}\\
	&HLR-M$^{2}$VC &\textbf{1.000(0.000)} &\textbf{1.000(0.000)} 	&\textbf{1.000(0.000)} &\textbf{1.000(0.000)} 	&\textbf{1.000(0.000)} 	&\textbf{1.000(0.000)} \\
	&COMSC &0.885(0.000) &0.883(0.000)	&0.977(0.000)	&0.881(0.000)	&0.872(0.000)	&0.940(0.000)\\
	&TBGL	&0.849(0.000)	&0.792(0.000)	&0.914(0.000)	&0.894(0.000)	&0.831(0.000)	&0.843(0.000)\\
	&TOMD-MVC 	&\textbf{1.000(0.000)} &\textbf{1.000(0.000)} 	&\textbf{1.000(0.000)} &\textbf{1.000(0.000)} 	&\textbf{1.000(0.000)} 	&\textbf{1.000(0.000)} \\
\hline
\end{tabular}
}
\end{center}
\end{table*}

\subsubsection{Compared with Tucker decomposition based methods}
In this section, we implement a number of experiments to evaluate the influence of applying higher-order form and O-minus structure on self-representation tensor $\mathcal{Z}$ in multi-view clustering. %Simultaneously, to demonstrate the significance of employing $\ominus$-like architecture for the core tensor from Tucker decomposition, the comparison between Tucker decomposition and TOMD is also presented. 
The primary procedure of clustering algorithm are same except for updating tensor $\mathcal{Z}$. Specifically, with 4-Dimensional Tucker decomposition based multi-view clustering model (4DT-MVC), we reshape $\mathcal{Z}$ into a 4th-order tensor and employs Tucker decomposition to capture its low-rank information while solving the problem with respect to $\mathcal{Z}$. As for the 3-Dimensional Tucker decomposition based MVC method (3DT-MVC), the representation tensor maintained 3rd-order. The average results after 10 runs of each experiment with six metrics are all presented in Table~\ref{tab:tdc}.

As results show, the improvements of 4DT-MVC are around $29.50\%$, $10.57\%$, $6.87\%$, $8.09\%$, $2.67\%$ about the average of six metrics over 3DT-MVC on Yale, MSRCV1, ExtendYaleB, ORL and Handwritten datasets, respectively. Especially for the Reuters data set, 4DT-MVC is more than twice as much as 3DT-MVC. The main reason may be that more low-rank information can be discovered in high-dimensional space. In addition, TOMD has a further $7.16\%$, $2.76\%$, $0.56\%$, $0.42\%$, $5.21\%$, $2.92\%$ development over 4DT-MVC on the same datasets, demonstrating the superiority of our proposed tensor network TOMD.

\begin{table*}[htbp]
\begin{center}
\caption{Clustering results on six datasets obtained by different tensor decomposition constraints.}
\label{tab:tdc}
\scalebox{0.8}{
\begin{tabular}{{cccccccccc}}
\hline		
Datasets &Method &F-score &Precision &Recall &NMI &AR &ACC
\\
\hline			
			&3DT-MVC 	&0.635 	&0.611 	&0.660 	&0.781 	&0.610 	&0.738  \\
Yale 			&4DT-MVC 	&0.842 	&0.825 	&0.861 	&0.913 	&0.832 	&0.927  \\
			&TOMD-MVC	&\textbf{0.916}	&\textbf{0.914}	&\textbf{0.918}	&\textbf{0.949}	&\textbf{0.910}	&\textbf{0.959} \\
\hline			
			&3DT-MVC	& 0.879	&0.866	&0.892	&0.869	&0.840	&0.943\\
MSRCV1  	&4DT-MVC  	& 0.972	&0.971	&0.973	&0.971	&0.967	&0.986\\
			&TOMD-MVC	& \textbf{1.000}	&\textbf{1.000}	&\textbf{1.000}	&\textbf{1.000}	&\textbf{1.000}	&\textbf{1.000}\\
\hline			
			&3DT-MVC 	& 0.909	 &0.908	&0.911	&0.905	&0.899	&0.954\\
ExtendYaleB 	&4DT-MVC  	& 0.975	&0.974	&0.975	&0.976	&0.973	&0.988\\
			&TOMD-MVC  	& \textbf{0.981} 	&\textbf{0.981}	&\textbf{0.981}	&\textbf{0.982}	&\textbf{0.979}	&\textbf{0.990}\\
\hline			
			&3DT-MVC	&0.908	&0.888	&0.928	&0.970	&0.905	&0.933\\
ORL 		&4DT-MVC  	&0.995	&0.994	&0.995	&0.998	&0.995  &0.998\\
			&TOMD-MVC 	&\textbf{1.000} 	&\textbf{1.000}	&\textbf{1.000}	&\textbf{1.000}	&\textbf{1.000}	&\textbf{1.000}\\
\hline			
			&3DT-MVC	& 0.445	&0.377	&0.543	&0.441	&0.310	&0.584\\
Reuters 		&4DT-MVC  	&0.922	&0.921	&0.924	&0.917	&0.907	&0.957\\
			&TOMD-MVC 	&\textbf{0.964} 	&\textbf{0.963}	&\textbf{0.966}	&\textbf{0.957}	&\textbf{0.957}	&\textbf{0.981}\\
\hline			
			&3DT-MVC	&0.945	&0.944	&0.945	&0.934	&0.939	&0.972\\
Handwritten  	&4DT-MVC  	&0.971	&0.971	&0.971	&0.963	&0.968	&0.986\\
			&TOMD-MVC 	&\textbf{1.000} &\textbf{1.000} &\textbf{1.000} &\textbf{1.000} 	&\textbf{1.000} 	&\textbf{1.000}\\
\hline
\end{tabular}
}
\end{center}
\end{table*}

\subsection{Model Discussion}
\subsubsection{Representation Visualization}
\begin{figure*}[htbp]
\centering
\subfloat[LMSC]{
\begin{minipage}[b]{.28\textwidth}
\centering
\includegraphics[scale=0.1]{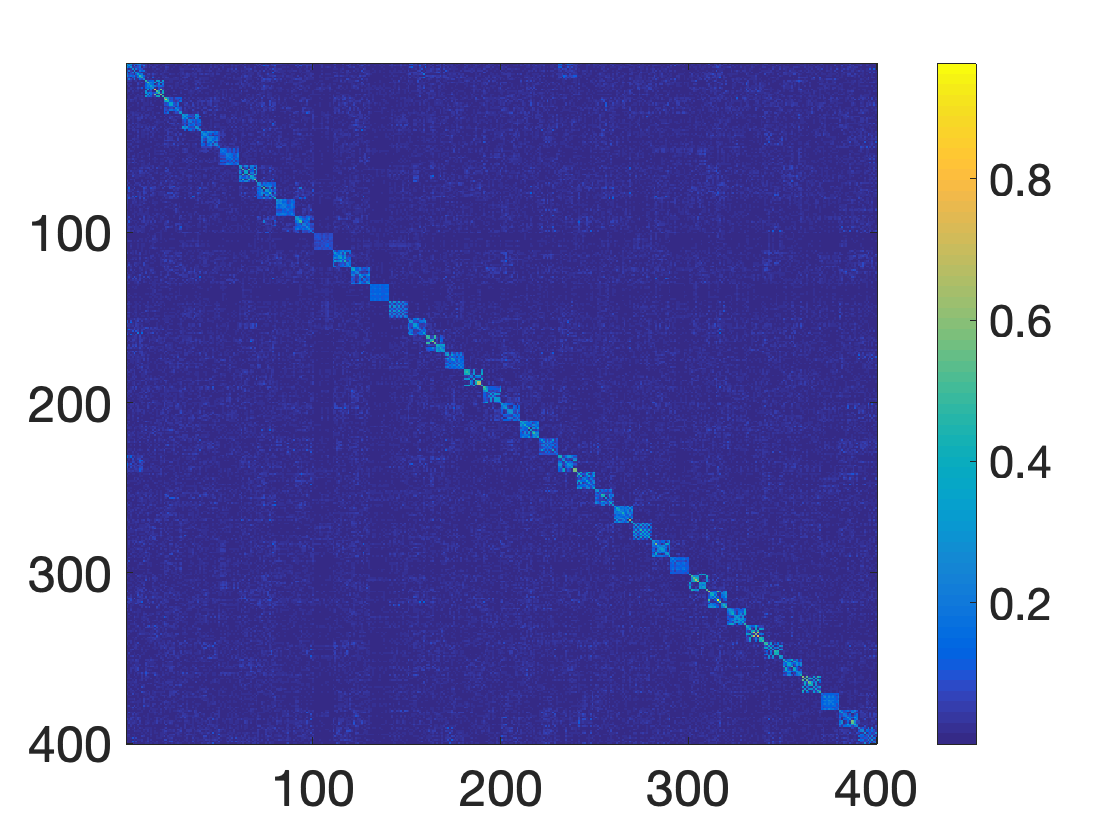}
\end{minipage}
}
\subfloat[t-SVD-MSC]{
\begin{minipage}[b]{.28\linewidth}
\centering
\includegraphics[scale=0.1]{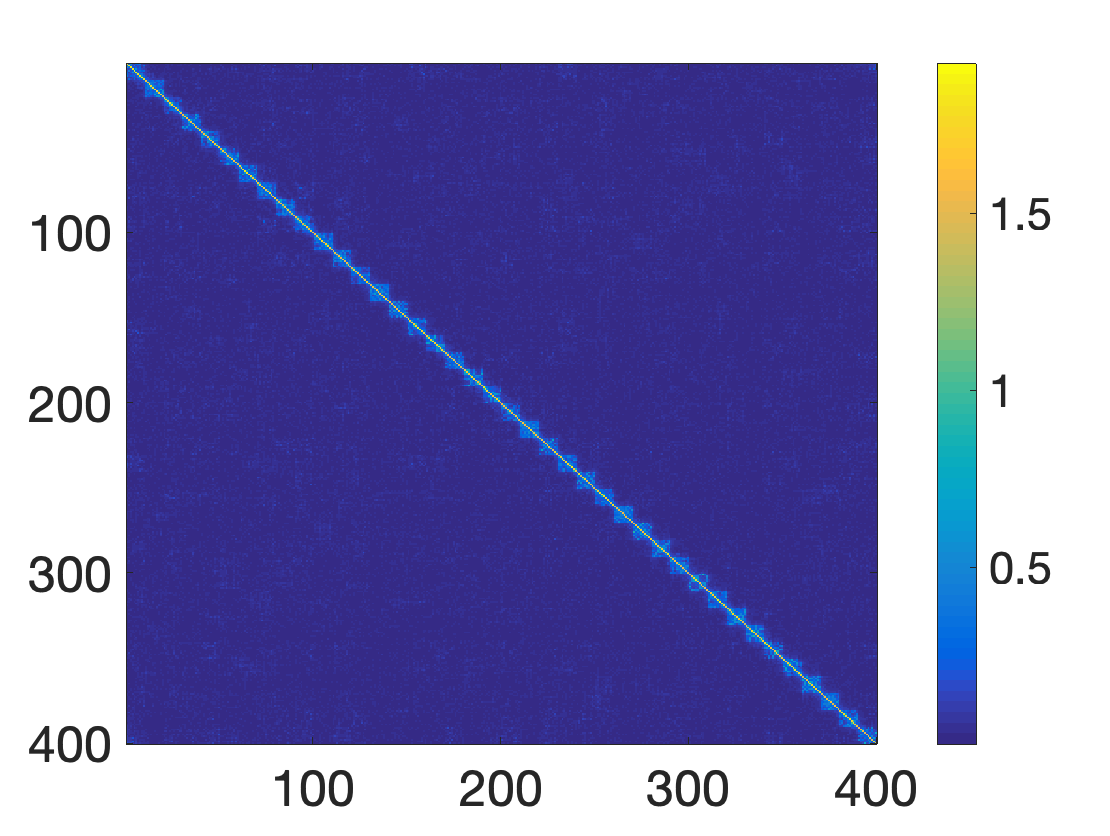}
\end{minipage}
}
\subfloat[TOMD-MVC]{
\begin{minipage}[b]{.28\linewidth}
\centering
\includegraphics[scale=0.1]{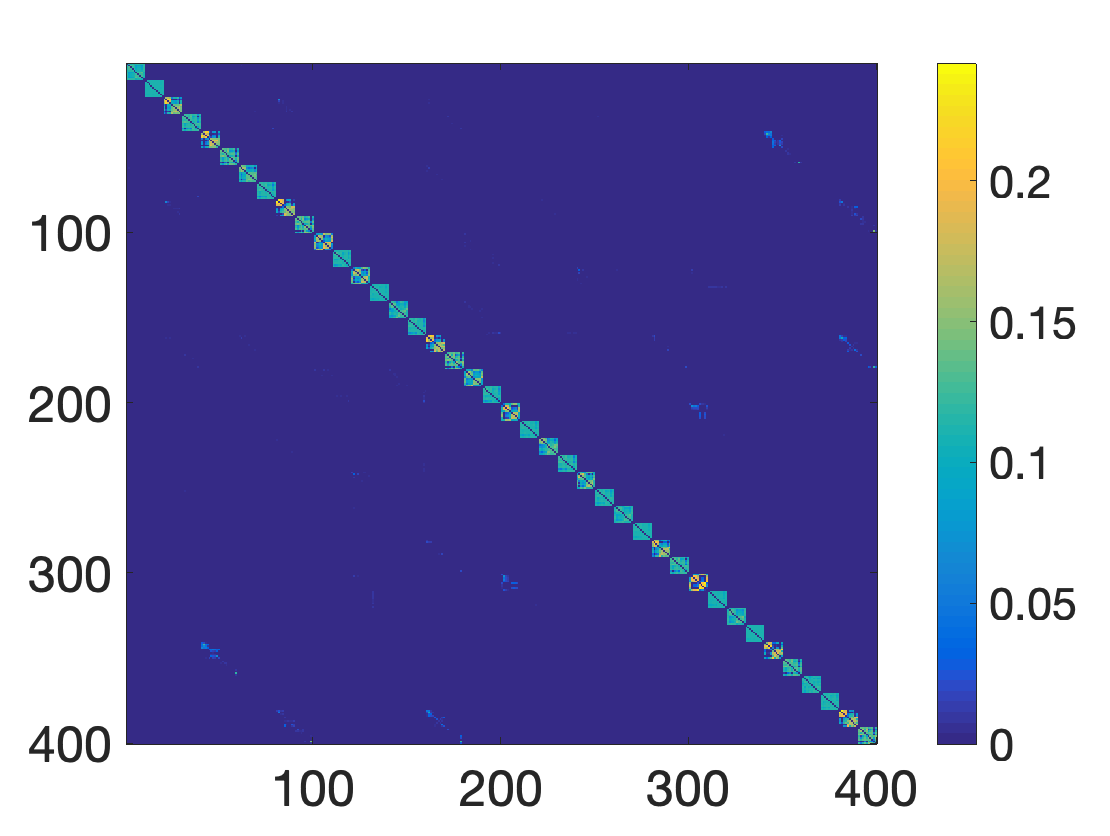}
\end{minipage}
}
\caption{Comparison of affinity matrix on ORL dataset by LMSC, t-SVD-MSC and TOMD-MVC.}
\label{fig:affinity}
\end{figure*}

Fig.~\ref{fig:affinity} visualizes the affinity matrices learned by different clustering methods. Due to the limitations of space, we only present the results of LMSC, t-SVD-MSC and TOMD-MVC on ORL dataset. Compared with the other three methods, our proposed TOMD-MVC is much better as its visualization matrix gives a more clear block-diagonal structure on ORL dataset. %This can also conclude from Table~\ref{tab:orl} that TOMD-MVC shows excellent clustering performance over the other methods.

% \begin{figure*}[htbp]
% \centering
% \includegraphics[scale=0.55]{Fig/viamatrix1.png}
% \caption{Comparison of affinity matrix on ORL dataset by LMSC, t-SVD-MSC and TOMD-MVC.}
% \label{fig:affinity}
% \end{figure*}

\subsubsection{Convergence}
\begin{figure*}[htbp]
\centering
\subfloat[3DT-MVC]{
\begin{minipage}[b]{.28\textwidth}
\centering
\includegraphics[scale=0.1]{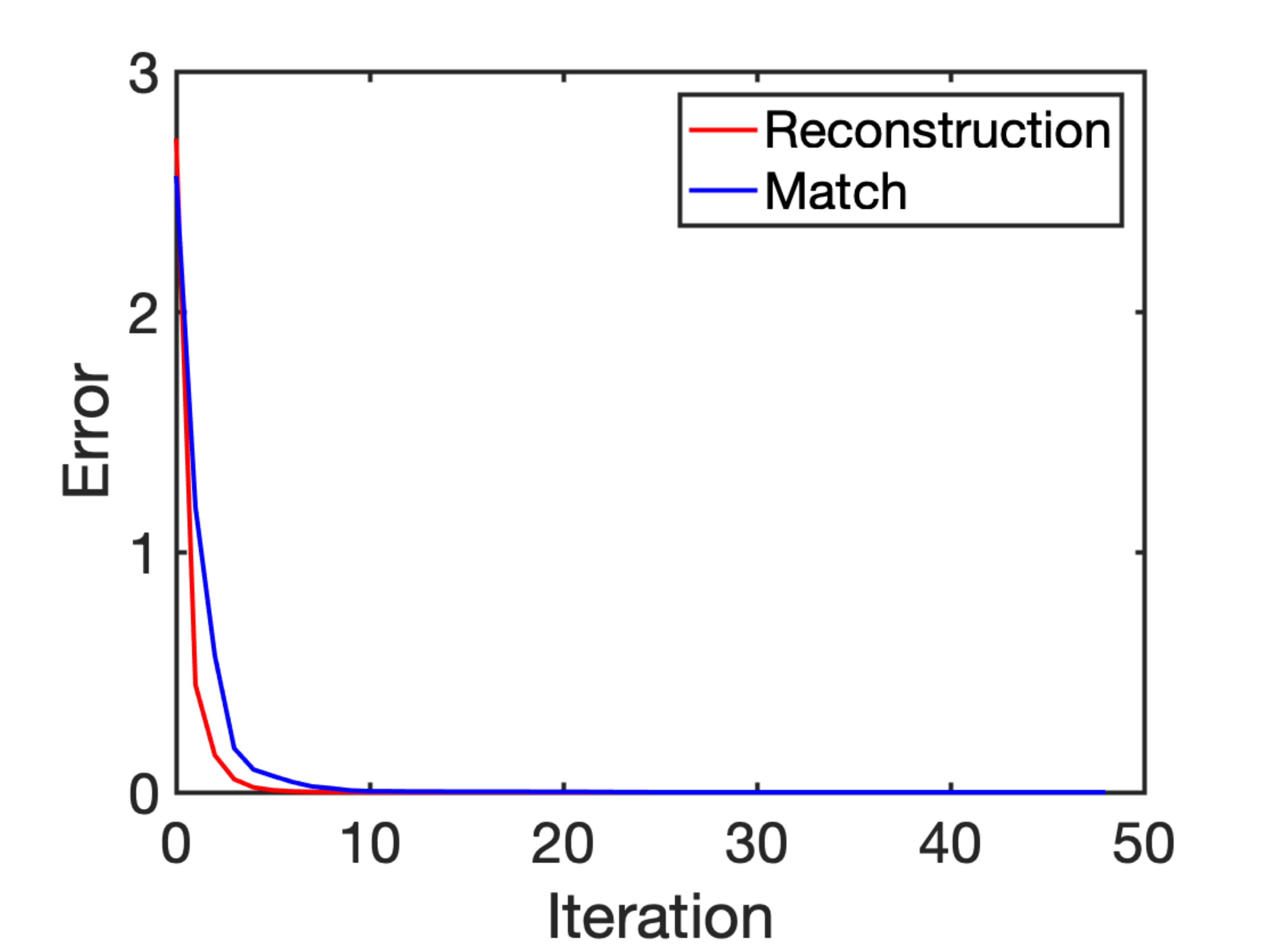}
\end{minipage}
}
\subfloat[4DT-MVC]{
\begin{minipage}[b]{.28\linewidth}
\centering
\includegraphics[scale=0.1]{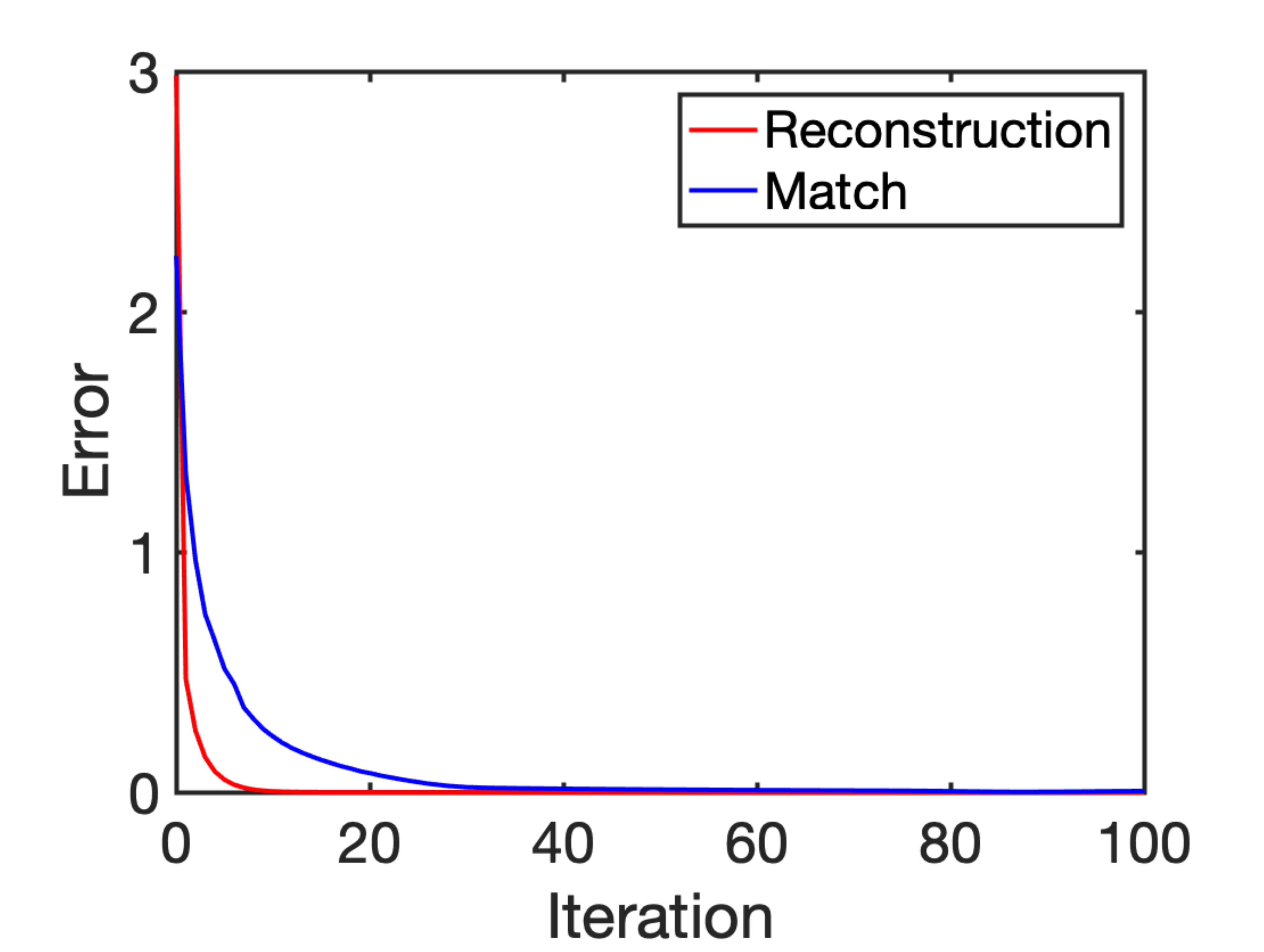}
\end{minipage}
}
\subfloat[TOMD-MVC]{
\begin{minipage}[b]{.28\linewidth}
\centering
\includegraphics[scale=0.1]{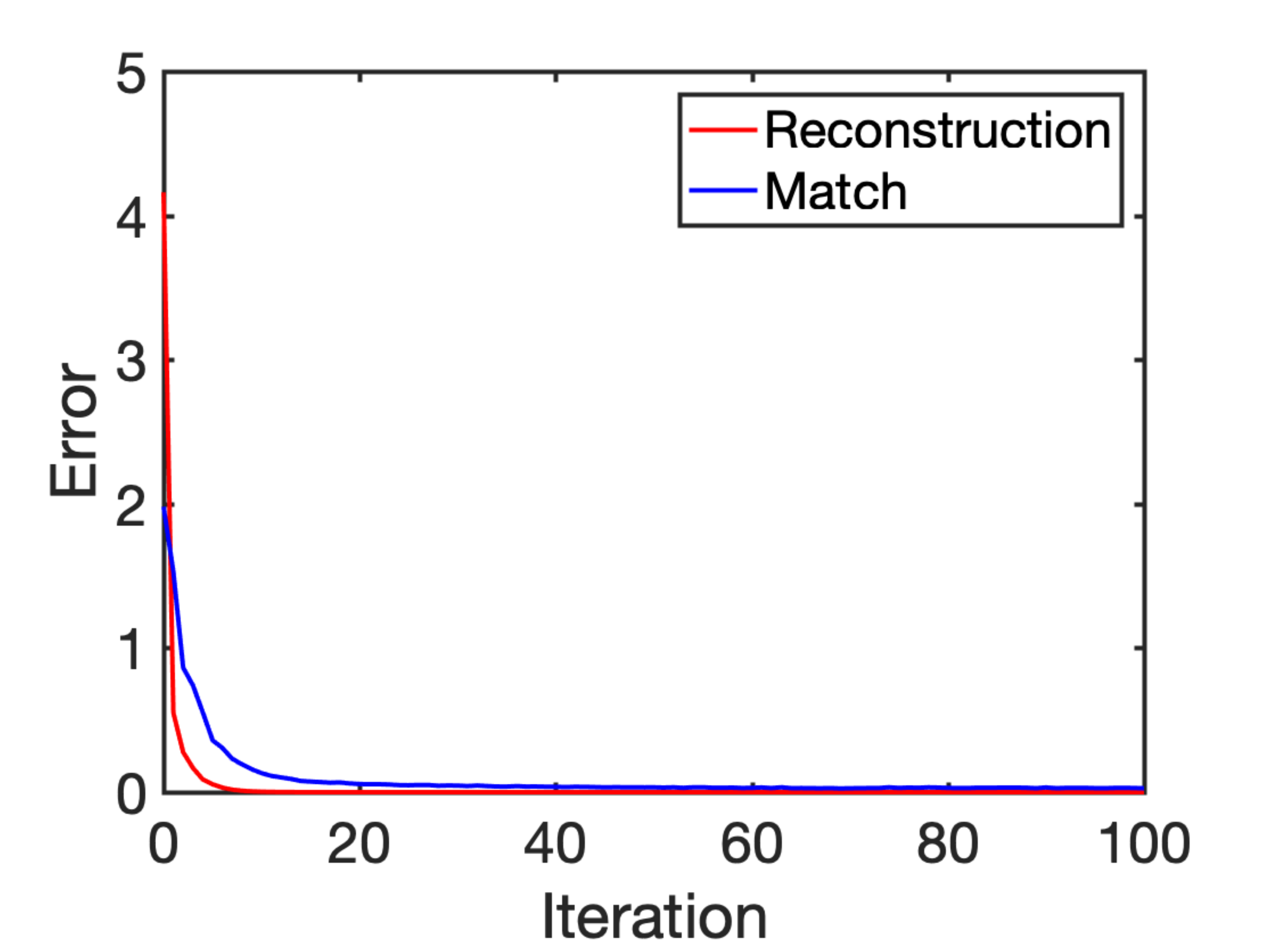}
\end{minipage}
}
\caption{The convergence curves on MSRCV1 dataset.}
\label{fig:convergence}
\end{figure*}

%In each iteration, we need to obtain the closed-form of $\mathcal{Z}$, $\mathcal{S}$, $\mathbf{E}$, and $\mathbf{M}$. As suggested in~\cite{boyd2011distributed,lin2011linearized}, though the convergence of ADMM with two blocks of variables has already been proved, its convergence properties with three or more blocks of variables have remained unclear. Accordingly, we present the empirical convergence of our proposed method in this subsection instead. 
In Fig.~\ref{fig:convergence}, we respectively plot the convergence curves of 3DT-MVC, 4DT-MVC and TOMD-MVC methods, where the X-axis is the number of iterations, and Y-axis is the value of $\text {Reconstruction Error}=(1/V) \sum_{v=1}^{V}\left\|\mathbf{X}_{v}-\mathbf{X}_{v} \mathbf{S}_{v}-\mathbf{E}_{v}\right\|_{\infty}$ and $\text {Match Error}=(1/V) \sum_{v=1}^{V}\left\|\mathbf{Z}_{v}-\mathbf{S}_{v}\right\|_{\infty}$. %(defined in~(\ref{eq:rematch})). 
As the results show, the 3DT-MVC convergences faster on the MSRCV1 dataset. However, its clustering performance does not perform well compared to the 4DT-MVC and TOMD-MVC. The convergence speed of the TOMD algorithm is similar to that of 4DT-MVC as presented in Fig.~\ref{fig:convergence}. Nonetheless, our proposed method has a further $2.76\%$ development over 4DT-MVC on MSRCV1 as illustrated in Table~\ref{tab:tdc}.  
%According to Fig.~\ref{fig:convergence}, it can be clearly seen that the reconstruction error and match error of TOMD decrease with the increasing of iteration number. 
Empirically, the number of optimization iteration is usually located within the range of $(100,150)$. 
% \begin{equation}
% \begin{aligned}
% \text {Reconstruction Error}&=\frac{1}{V} \sum_{v=1}^{V}\left\|\mathbf{X}_{v}-\mathbf{X}_{v} \mathbf{S}_{v}-\mathbf{E}_{v}\right\|_{\infty}\\
% \text {Match Error}&=\frac{1}{V} \sum_{v=1}^{V}\left\|\mathbf{Z}_{v}-\mathbf{S}_{v}\right\|_{\infty}.
% \end{aligned}
% \label{eq:rematch}
% \end{equation}

\subsubsection{Parameter Sensitivity Analysis}
\begin{figure}[htbp]
\centering
\includegraphics[scale=0.17]{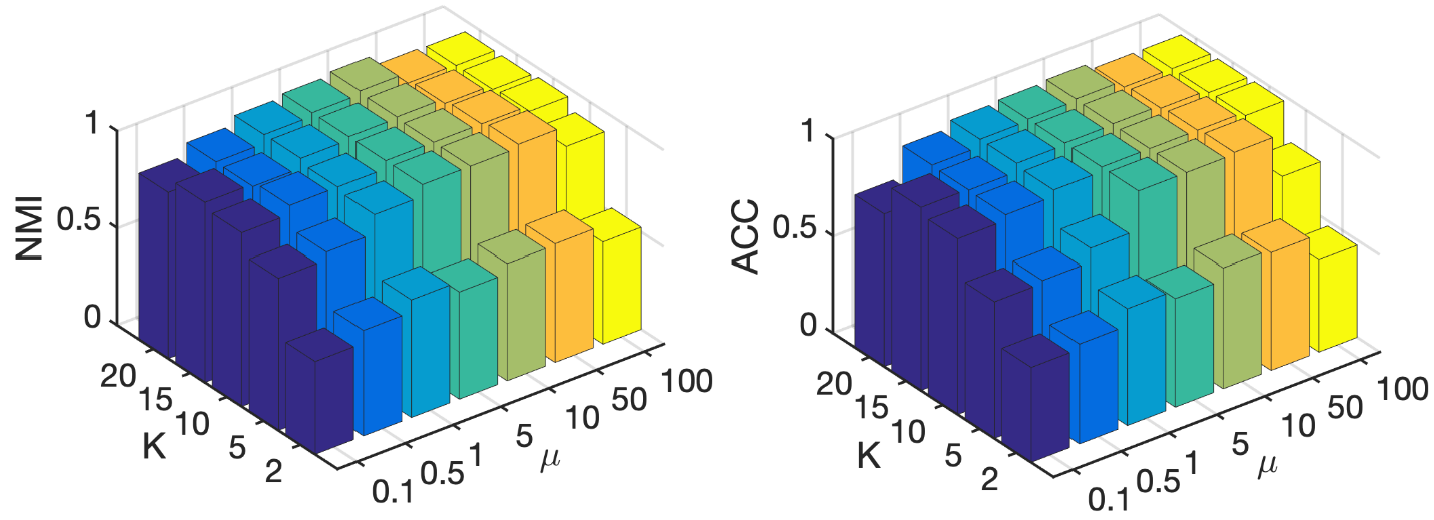}
\caption{Parameters tuning ($\mu$ and $K$) in terms of ACC and NMI on MSRCV1.}
\label{fig:para}
\end{figure}

As suggested in~\cite{chen2021low}, the parameter $\lambda$ can be determined by the number of adaptive neighbors $K$. Therefore, there are three free parameters in our model: balanced parameter $\mu$, the number of adaptive neighbors $K$ and tensor ranks $(R_{1},\cdots,R_{4},D_{1},\cdots,D_{6})$. Due to the limitations of the paper, we only show the results of ACC and NMI on Yale dataset by the combination of different $\mu$ and $K$ in Fig.~\ref{fig:para}. We can see that the performance of TOMD-MVC is relatively stable when $\mu=[5,50]$ and $K=[5,15]$ on Yale dataset. Besides, the evaluation results on Yale dataset have been presented in Table~\ref{tab:ranks} with different rank tuning. The results show that TOMD-MVC can achieve promising performance while the pre-defined ranks are set as $(30,15,11,V,4,4,4,4,4,4)$, where $V$ is the number of views.
%As suggested in~\cite{chen2021low}, the parameter $\lambda$ can be determined by the number of adaptive neighbors $K$. Therefore, there are three free parameters in our model: balanced parameter $\mu$, the number of adaptive neighbors $K$ and tensor ranks. Due to the limitations of the paper, we only show the results of ACC and NMI on Yale dataset by the combination of different $\mu$ and $K$ in Fig.~\ref{fig:para}. We can see that the performance of TOMD-MVC is relatively stable when $\mu=[5,50]$ and $K=[5,15]$ on Yale dataset. Besides, the evaluation results on Yale dataset have been presented in Table~\ref{tab:ranks} with different rank tuning. The results show that TOMD-MVC can achieve promising performance while the pre-defined ranks are set as $(30,15,11,V,4,4,4,4,4,4)$, where $V$ is the number of views.

\begin{table*}[htbp]
\begin{center}
\caption{Clustering results of TOMD-MVC with different ranks on Yale dataset.}
\label{tab:ranks}
\scalebox{0.6}{
\begin{tabular}{{ccccccccc}}
\hline		
 $(R_1,\cdots,R_4,D_1,\cdots,D_6)$  &F-score &Precision &Recall &NMI &AR &ACC
\\
\hline
	$(10,5,5,V,2,2,2,2,2,2)$ 	&0.626(0.018)	&0.595(0.016)	&0.660(0.021)	&0.812(0.010)	&0.600(0.020)	&0.723(0.029) \\
	$(10,10,5,V,2,2,2,2,2,2)$ 	&0.624(0.019)	&0.594(0.016)	&0.657(0.023)	&0.808(0.012)	&0.598(0.020)	&0.730(0.020) \\
	$(10,5,10,V,2,2,2,2,2,2)$ 	&0.623(0.008)	&0.583(0.007)	&0.670(0.011)	&0.813(0.005)	&0.597(0.009)	&0.707(0.020) \\
	$(20,5,5,V,2,2,2,2,2,2)$	&0.630(0.017)	&0.598(0.015)	&0.666(0.020)	&0.813(0.010)	&0.605(0.019)	&0.724(0.034) \\
	$(10,5,5,V,4,4,4,4,4,4)$	&0.490(0.032)	&0.464(0.035)	&0.520(0.030)	&0.711(0.021)	&0.455(0.035)	&0.652(0.034) \\
 	$(20,10,10,V,4,4,4,4,4,4)$ &0.844(0.022)	&0.835(0.022)	&0.852(0.022)	&0.910(0.014)	&0.833(0.023)	&0.921(0.012) \\
	$(20,15,10,V,4,4,4,4,4,4)$	&0.843(0.012)	&0.833(0.015)	&0.854(0.010)	&0.909(0.006)	&0.833(0.013)	&0.922(0.006) \\
	$(20,10,11,V,4,4,4,4,4,4)$ &0.887(0.015)	&0.884(0.015)	&0.891(0.015)	&0.932(0.010)	&0.880(0.016)	&0.944(0.008) \\
	$(30,10,10,V,4,4,4,4,4,4)$ &0.858(0.012)	&0.849(0.014)	&0.866(0.010)	&0.918(0.006)	 &0.848(0.013)	&0.929(0.006) \\
	$(30,15,11,V,6,6,6,6,6,6)$	&0.637(0.007)	&0.613(0.010)	&0.663(0.005)	&0.782(0.004)	&0.613(0.008)	&0.738(0.006) \\
	$(30,15,11,V,4,4,4,4,4,4)$	&\textbf{0.916(0.013)}	&\textbf{0.914(0.014)}	&\textbf{0.918(0.013)}	&\textbf{0.949(0.008)}	&\textbf{0.910(0.014)}	&\textbf{0.959(0.006)}\\
	$(40,15,11,V,4,4,4,4,4,4)$  &0.896(0.008)	&0.892(0.009)	&0.899(0.008)	&0.937(0.005)	&0.889(0.009)	&0.948(0.004)  \\
\hline
\end{tabular}
}
\end{center}
\end{table*}

\section{Conclusion}
\label{sec:7}
In this paper, we propose a novel tensor network called TOMD to considerably capture the low-rank property of representation tensor and fully explore the latent correlations across different views in clustering. In addition, we drive a detailed algorithm to solve the optimization problem. Accordingly, TOMD-MVC has been developed and its superiority performance has been demonstrated by extensive experimental results on six real-world datasets. 

%Considering that the optimal rank selection is essential in tensor decomposition, we would like to focus on introducing an automatically selecting ranks strategy with lower computational complexity in the future. %which probably further reduces its storage cost.
%Considering that the number of iterations mainly influences the TOMD-ALS algorithm and the contractions of tensor networks with loops will cost high complexity, we would like to focus on fast algorithms in the future. Secondly, the optimal rank selection is essential in tensor decomposition. Therefore, introducing an automatically selecting ranks strategy could be another direction. %which probably further reduces its storage cost.

\section*{Acknowledgment}

This work was supported by the National Natural Science Foundation of China (NSFC) under Grant 62171088, Grant 6220106011 and Grant U19A2052.

%\section*{References}
\bibliographystyle{plain}
\bibliography{citations}

\begin{thebibliography}{10}

\bibitem{adali2022reproducibility}
T{\"u}lay Adali, Furkan Kantar, Mohammad Abu Baker~Siddique Akhonda, Stephen
  Strother, Vince~D Calhoun, and Evrim Acar.
\newblock Reproducibility in matrix and tensor decompositions: Focus on model
  match, interpretability, and uniqueness.
\newblock {\em IEEE Signal Processing Magazine}, 39(4):8--24, 2022.

\bibitem{bengua2017efficient}
Johann~A Bengua, Ho~N Phien, Hoang~Duong Tuan, and Minh~N Do.
\newblock Efficient tensor completion for color image and video recovery:
  Low-rank tensor train.
\newblock {\em IEEE Transactions on Image Processing}, 26(5):2466--2479, 2017.

\bibitem{billo2019two}
M~Bill{\'o}, F~Fucito, GP~Korchemsky, A~Lerda, and JF~Morales.
\newblock Two-point correlators in non-conformal $\mathcal{N}$= 2 gauge
  theories.
\newblock {\em Journal of High Energy Physics}, 2019(5):1--56, 2019.

\bibitem{carroll1970analysis}
J~Douglas Carroll and Jih-Jie Chang.
\newblock Analysis of individual differences in multidimensional scaling via an
  n-way generalization of “eckart-young” decomposition.
\newblock {\em Psychometrika}, 35(3):283--319, 1970.

\bibitem{chen2021tensor}
Hongyang Chen, Fauzia Ahmad, Sergiy Vorobyov, and Fatih Porikli.
\newblock Tensor decompositions in wireless communications and mimo radar.
\newblock {\em IEEE Journal of Selected Topics in Signal Processing},
  15(3):438--453, 2021.

\bibitem{chen2021smoothed}
Peng Chen, Liang Liu, Zhengrui Ma, and Zhao Kang.
\newblock Smoothed multi-view subspace clustering.
\newblock In {\em International Conference on Neural Computing for Advanced
  Applications}, pages 128--140. Springer, 2021.

\bibitem{chen2021low}
Yongyong Chen, Xiaolin Xiao, Chong Peng, Guangming Lu, and Yicong Zhou.
\newblock Low-rank tensor graph learning for multi-view subspace clustering.
\newblock {\em IEEE Transactions on Circuits and Systems for Video Technology},
  2021.

\bibitem{chen2021hierarchical}
Zefeng Chen, Guoxu Zhou, and Qibin Zhao.
\newblock Hierarchical factorization strategy for high-order tensor and
  application for data completion.
\newblock {\em IEEE Signal Processing Letters}, 2021.

\bibitem{cichocki2016low}
Andrzej Cichocki, Namgil Lee, Ivan~V Oseledets, A-H Phan, Qibin Zhao, and
  D~Mandic.
\newblock Low-rank tensor networks for dimensionality reduction and large-scale
  optimization problems: Perspectives and challenges part 1.
\newblock {\em Foundations and Trends in Machine Learning}, 9(4-5):249--429,
  2016.

\bibitem{feng2021multi}
Lanlan Feng, Ce~Zhu, and Yipeng Liu.
\newblock Multi-mode tensor singular value decomposition for low-rank image
  recovery.
\newblock In {\em International Conference on Image and Graphics}, pages
  238--249. Springer, 2021.

\bibitem{hartigan1979algorithm}
John~A Hartigan and Manchek~A Wong.
\newblock Algorithm as 136: A k-means clustering algorithm.
\newblock {\em Journal of The Royal Statistical Society. Series C (Applied
  Statistics)}, 28(1):100--108, 1979.

\bibitem{he2018self}
Lifang He, Chun-Ta Lu, Yong Chen, Jiawei Zhang, Linlin Shen, S~Yu Philip, and
  Fei Wang.
\newblock A self-organizing tensor architecture for multi-view clustering.
\newblock In {\em 2018 IEEE International Conference on Data Mining (ICDM)},
  pages 1007--1012. IEEE, 2018.

\bibitem{hou2017multi}
Chenping Hou, Feiping Nie, Hong Tao, and Dongyun Yi.
\newblock Multi-view unsupervised feature selection with adaptive similarity
  and view weight.
\newblock {\em IEEE Transactions on Knowledge and Data Engineering},
  29(9):1998--2011, 2017.

\bibitem{huang2019ultra}
Dong Huang, Chang-Dong Wang, Jian-Sheng Wu, Jian-Huang Lai, and Chee-Keong
  Kwoh.
\newblock Ultra-scalable spectral clustering and ensemble clustering.
\newblock {\em IEEE Transactions on Knowledge and Data Engineering},
  32(6):1212--1226, 2019.

\bibitem{ji2019survey}
Yuwang Ji, Qiang Wang, Xuan Li, and Jie Liu.
\newblock A survey on tensor techniques and applications in machine learning.
\newblock {\em IEEE Access}, 7:162950--162990, 2019.

\bibitem{kaliyar2021deepfake}
Rohit~Kumar Kaliyar, Anurag Goswami, and Pratik Narang.
\newblock Deepfake: improving fake news detection using tensor
  decomposition-based deep neural network.
\newblock {\em The Journal of Supercomputing}, 77(2):1015--1037, 2021.

\bibitem{kang2019robust}
Zhao Kang, Haiqi Pan, Steven~CH Hoi, and Zenglin Xu.
\newblock Robust graph learning from noisy data.
\newblock {\em IEEE Transactions on Cybernetics}, 50(5):1833--1843, 2019.

\bibitem{kang2020multi}
Zhao Kang, Guoxin Shi, Shudong Huang, Wenyu Chen, Xiaorong Pu, Joey~Tianyi
  Zhou, and Zenglin Xu.
\newblock Multi-graph fusion for multi-view spectral clustering.
\newblock {\em Knowledge-Based Systems}, 189:105102, 2020.

\bibitem{kiers2000towards}
Henk~AL Kiers.
\newblock Towards a standardized notation and terminology in multiway analysis.
\newblock {\em Journal of Chemometrics: A Journal of the Chemometrics Society},
  14(3):105--122, 2000.

\bibitem{kilmer2011factorization}
Misha~E Kilmer and Carla~D Martin.
\newblock Factorization strategies for third-order tensors.
\newblock {\em Linear Algebra and its Applications}, 435(3):641--658, 2011.

\bibitem{kilmer2008third}
Misha~E Kilmer, Carla~D Martin, and Lisa Perrone.
\newblock A third-order generalization of the matrix svd as a product of
  third-order tensors.
\newblock {\em Tufts University, Department of Computer Science, Tech. Rep.
  TR-2008-4}, 2008.

\bibitem{kolda2009tensor}
Tamara~G Kolda and Brett~W Bader.
\newblock Tensor decompositions and applications.
\newblock {\em SIAM Review}, 51(3):455--500, 2009.

\bibitem{kroonenberg1980principal}
Pieter~M Kroonenberg and Jan De~Leeuw.
\newblock Principal component analysis of three-mode data by means of
  alternating least squares algorithms.
\newblock {\em Psychometrika}, 45(1):69--97, 1980.

\bibitem{li2020multiview}
Xuelong Li, Han Zhang, Rong Wang, and Feiping Nie.
\newblock Multiview clustering: A scalable and parameter-free bipartite graph
  fusion method.
\newblock {\em IEEE Transactions on Pattern Analysis and Machine Intelligence},
  44(1):330--344, 2020.

\bibitem{lin2018multi}
Kun-Yu Lin, Ling Huang, Chang-Dong Wang, and Hong-Yang Chao.
\newblock Multi-view proximity learning for clustering.
\newblock In {\em International Conference on Database Systems for Advanced
  Applications}, pages 407--423. Springer, 2018.

\bibitem{liu2012robust}
Guangcan Liu, Zhouchen Lin, Shuicheng Yan, Ju~Sun, Yong Yu, and Yi~Ma.
\newblock Robust recovery of subspace structures by low-rank representation.
\newblock {\em IEEE Transactions on Pattern Analysis and Machine Intelligence},
  35(1):171--184, 2012.

\bibitem{liu2020smooth}
Jiani Liu, Ce~Zhu, and Yipeng Liu.
\newblock Smooth compact tensor ring regression.
\newblock {\em IEEE Transactions on Knowledge and Data Engineering}, 2020.

\bibitem{liu2021multiview}
Jiyuan Liu, Xinwang Liu, Yuexiang Yang, Xifeng Guo, Marius Kloft, and
  Liangzhong He.
\newblock Multiview subspace clustering via co-training robust data
  representation.
\newblock {\em IEEE Transactions on Neural Networks and Learning Systems},
  2021.

\bibitem{liu2022efficient}
Suyuan Liu, Siwei Wang, Pei Zhang, Kai Xu, Xinwang Liu, Changwang Zhang, and
  Feng Gao.
\newblock Efficient one-pass multi-view subspace clustering with consensus
  anchors.
\newblock In {\em Proceedings of the AAAI Conference on Artificial
  Intelligence}, volume~36, pages 7576--7584, 2022.

\bibitem{liu2021tensor}
Yipeng Liu, Jiani Liu, Zhen Long, and Zhu Ce.
\newblock {\em Tensor Computation for Data Analysis}.
\newblock Springer International Publishing, 2021.

\bibitem{liu2020generalizing}
Yu~Liu, Quanming Yao, and Yong Li.
\newblock Generalizing tensor decomposition for n-ary relational knowledge
  bases.
\newblock In {\em Proceedings of The Web Conference 2020}, pages 1104--1114,
  2020.

\bibitem{long2019low}
Zhen Long, Yipeng Liu, Longxi Chen, and Ce~Zhu.
\newblock Low rank tensor completion for multiway visual data.
\newblock {\em Signal processing}, 155:301--316, 2019.

\bibitem{long2021bayesian}
Zhen Long, Ce~Zhu, Jiani Liu, and Yipeng Liu.
\newblock Bayesian low rank tensor ring for image recovery.
\newblock {\em IEEE Transactions on Image Processing}, 30:3568--3580, 2021.

\bibitem{ng2002spectral}
Andrew~Y Ng, Michael~I Jordan, and Yair Weiss.
\newblock On spectral clustering: Analysis and an algorithm.
\newblock In {\em Advances in Neural Information Processing Systems}, pages
  849--856, 2002.

\bibitem{nie2017multi}
Feiping Nie, Guohao Cai, and Xuelong Li.
\newblock Multi-view clustering and semi-supervised classification with
  adaptive neighbours.
\newblock In {\em Thirty-first AAAI Conference on Artificial Intelligence},
  2017.

\bibitem{nie2016parameter}
Feiping Nie, Jing Li, and Xuelong Li.
\newblock Parameter-free auto-weighted multiple graph learning: a framework for
  multiview clustering and semi-supervised classification.
\newblock In {\em International Joint Conference on Artificial Intelligence},
  pages 1881--1887, 2016.

\bibitem{nie2017self}
Feiping Nie, Jing Li, and Xuelong Li.
\newblock Self-weighted multiview clustering with multiple graphs.
\newblock In {\em International Joint Conference on Artificial Intelligence},
  pages 2564--2570, 2017.

\bibitem{ran2020tensor}
Shi-Ju Ran, Emanuele Tirrito, Cheng Peng, Xi~Chen, Luca Tagliacozzo, Gang Su,
  and Maciej Lewenstein.
\newblock {\em Tensor Network Contractions: Methods and Applications to Quantum
  Many-Body Systems}.
\newblock Springer Nature, 2020.

\bibitem{schutze2008introduction}
Hinrich Sch{\"u}tze, Christopher~D Manning, and Prabhakar Raghavan.
\newblock {\em Introduction to information retrieval}, volume~39.
\newblock Cambridge University Press Cambridge, 2008.

\bibitem{sui2019sparse}
Yao Sui, Guanghui Wang, and Li~Zhang.
\newblock Sparse subspace clustering via low-rank structure propagation.
\newblock {\em Pattern Recognition}, 95:261--271, 2019.

\bibitem{tucker1966some}
Ledyard~R Tucker.
\newblock Some mathematical notes on three-mode factor analysis.
\newblock {\em Psychometrika}, 31(3):279--311, 1966.

\bibitem{tucker1964extension}
Ledyard~R Tucker et~al.
\newblock The extension of factor analysis to three-dimensional matrices.
\newblock {\em Contributions to mathematical psychology}, 110119, 1964.

\bibitem{wang2019gmc}
Hao Wang, Yan Yang, and Bing Liu.
\newblock Gmc: Graph-based multi-view clustering.
\newblock {\em IEEE Transactions on Knowledge and Data Engineering},
  32(6):1116--1129, 2019.

\bibitem{wang2017exclusivity}
Xiaobo Wang, Xiaojie Guo, Zhen Lei, Changqing Zhang, and Stan~Z Li.
\newblock Exclusivity-consistency regularized multi-view subspace clustering.
\newblock In {\em Proceedings of the IEEE Conference on Computer Vision and
  Pattern Recognition}, pages 923--931, 2017.

\bibitem{wu2019essential}
Jianlong Wu, Zhouchen Lin, and Hongbin Zha.
\newblock Essential tensor learning for multi-view spectral clustering.
\newblock {\em IEEE Transactions on Image Processing}, 28(12):5910--5922, 2019.

\bibitem{wu2020unified}
Jianlong Wu, Xingyu Xie, Liqiang Nie, Zhouchen Lin, and Hongbin Zha.
\newblock Unified graph and low-rank tensor learning for multi-view clustering.
\newblock In {\em Proceedings of the AAAI conference on artificial
  intelligence}, volume~34, pages 6388--6395, 2020.

\bibitem{xia2022tensorized}
Wei Xia, Quanxue Gao, Qianqian Wang, Xinbo Gao, Chris Ding, and Dacheng Tao.
\newblock Tensorized bipartite graph learning for multi-view clustering.
\newblock {\em IEEE Transactions on Pattern Analysis and Machine Intelligence},
  2022.

\bibitem{xie2018unifying}
Yuan Xie, Dacheng Tao, Wensheng Zhang, Yan Liu, Lei Zhang, and Yanyun Qu.
\newblock On unifying multi-view self-representations for clustering by tensor
  multi-rank minimization.
\newblock {\em International Journal of Computer Vision}, 126(11):1157--1179,
  2018.

\bibitem{xie2020hyper}
Yuan Xie, Wensheng Zhang, Yanyun Qu, Longquan Dai, and Dacheng Tao.
\newblock Hyper-laplacian regularized multilinear multiview
  self-representations for clustering and semisupervised learning.
\newblock {\em IEEE transactions on cybernetics}, 50(2):572--586, 2020.

\bibitem{yang2018multi}
Yan Yang and Hao Wang.
\newblock Multi-view clustering: A survey.
\newblock {\em Big Data Mining and Analytics}, 1(2):83--107, 2018.

\bibitem{yu2020graph}
Yuyuan Yu, Guoxu Zhou, Ning Zheng, Shengli Xie, and Qibin Zhao.
\newblock Graph regularized nonnegative tensor ring decomposition for multiway
  representation learning.
\newblock {\em arXiv preprint arXiv:2010.05657}, 2020.

\bibitem{yuan2018higher}
Longhao Yuan, Jianting Cao, Xuyang Zhao, Qiang Wu, and Qibin Zhao.
\newblock Higher-dimension tensor completion via low-rank tensor ring
  decomposition.
\newblock In {\em 2018 Asia-Pacific Signal and Information Processing
  Association Annual Summit and Conference (APSIPA ASC)}, pages 1071--1076.
  IEEE, 2018.

\bibitem{zhang2015low}
Changqing Zhang, Huazhu Fu, Si~Liu, Guangcan Liu, and Xiaochun Cao.
\newblock Low-rank tensor constrained multiview subspace clustering.
\newblock In {\em Proceedings of the IEEE International Conference on Computer
  Vision}, pages 1582--1590, 2015.

\bibitem{zhang2020tensorized}
Changqing Zhang, Huazhu Fu, Jing Wang, Wen Li, Xiaochun Cao, and Qinghua Hu.
\newblock Tensorized multi-view subspace representation learning.
\newblock {\em International Journal of Computer Vision}, 128(8):2344--2361,
  2020.

\bibitem{zhang2017latent}
Changqing Zhang, Qinghua Hu, Huazhu Fu, Pengfei Zhu, and Xiaochun Cao.
\newblock Latent multi-view subspace clustering.
\newblock In {\em Proceedings of the IEEE Conference on Computer Vision and
  Pattern Recognition}, pages 4279--4287, 2017.

\bibitem{zhang2021joint}
Guang-Yu Zhang, Yu-Ren Zhou, Chang-Dong Wang, Dong Huang, and Xiao-Yu He.
\newblock Joint representation learning for multi-view subspace clustering.
\newblock {\em Expert Systems with Applications}, 166:113913, 2021.

\bibitem{zhao2016tensor}
Qibin Zhao, Guoxu Zhou, Shengli Xie, Liqing Zhang, and Andrzej Cichocki.
\newblock Tensor ring decomposition.
\newblock {\em arXiv preprint arXiv:1606.05535}, 2016.

\bibitem{zheng2022multi}
Qinghai Zheng and Jihua Zhu.
\newblock Multi-view subspace clustering with view correlations via low-rank
  tensor learning.
\newblock {\em Computers and Electrical Engineering}, 100:107939, 2022.

\bibitem{zheng2020constrained}
Qinghai Zheng, Jihua Zhu, Zhiqiang Tian, Zhongyu Li, Shanmin Pang, and Xiuyi
  Jia.
\newblock Constrained bilinear factorization multi-view subspace clustering.
\newblock {\em Knowledge-Based Systems}, 194:105--514, 2020.

\bibitem{zhong2003unified}
Shi Zhong and Joydeep Ghosh.
\newblock A unified framework for model-based clustering.
\newblock {\em The Journal of Machine Learning Research}, 4:1001--1037, 2003.

\bibitem{zhuge2017unsupervised}
Wenzhang Zhuge, Feiping Nie, Chenping Hou, and Dongyun Yi.
\newblock Unsupervised single and multiple views feature extraction with
  structured graph.
\newblock {\em IEEE Transactions on Knowledge and Data Engineering},
  29(10):2347--2359, 2017.

\end{thebibliography}

\end{document}